\documentclass[pdflatex,sn-mathphys-num]{sn-jnl}% Math and Physical Sciences Numbered Reference Style
\usepackage{graphicx}%
\usepackage{multirow}%
\usepackage{amsmath,amssymb,amsfonts}%
\usepackage{amsthm}%
\usepackage[title]{appendix}%
\usepackage{xcolor}%
\usepackage{textcomp}%
\usepackage{manyfoot}%
\usepackage{booktabs}%
\usepackage{algorithm}%
\usepackage{algorithmicx}%
\usepackage{algpseudocode}%
\usepackage{listings}%
\usepackage{tabularx}%
\usepackage{placeins}%
\usepackage{float}%
\theoremstyle{thmstyleone}%
\theoremstyle{thmstyletwo}%

\theoremstyle{thmstylethree}%

\begin{document}

\title{Generalization Under Scrutiny: Cross-Domain Detection Progresses, Pitfalls, and Persistent Challenges}

\author*[1]{\fnm{Saniya M.} \sur{Deshmukh}}\email{saniya.deshmukh@ubi.pt}

\author[1]{\fnm{Kailash A.} \sur{Hambarde}}\email{kailash.hambarde@ubi.pt}

\author[1]{\fnm{Hugo} \sur{Proen\c{c}a}}\email{hugomcp@ubi.pt}

\affil*[1]{\orgdiv{IT: Instituto de Telecomunica\c{c}\~oes}, \orgname{University of Beira Interior}, \orgaddress{\city{Covilh\~a}, \country{Portugal}}}

%%==================================%%
%%==================================%%

\abstract{Object detection models trained on a source domain often exhibit significant performance degradation when deployed in unseen target domains, due to various kinds of variations, such as sensing conditions, environments and data distributions. Hence, regardless the recent breakthrough advances in deep learning-based detection technology, cross-domain object detection (CDOD) remains a critical research area.
Moreover, the existing literature remains fragmented, lacking a unified perspective on the structural challenges underlying domain shift and the effectiveness of adaptation strategies.
This survey provides a comprehensive and systematic analysis of CDOD. We start upon a problem formulation that highlights the multi-stage nature of object detection under domain shift. Then, we  organize the existing methods through a conceptual taxonomy that categorizes approaches based on adaptation paradigms, modeling assumptions, and pipeline components. Furthermore, we analyze how domain shift propagates across detection stages and discuss why adaptation in object detection is inherently more complex than in classification. In addition, we review commonly used datasets, evaluation protocols, and benchmarking practices. Finally, we identify the key challenges and outline promising future research directions. Cohesively, this survey aims to provide a unified framework for understanding CDOD and to guide the development of more robust detection systems.}

%%================================%%




\maketitle

\section{Introduction}
\label{sec:introduction}

\begin{figure}[H]
\centering
\includegraphics[width=0.5\columnwidth]{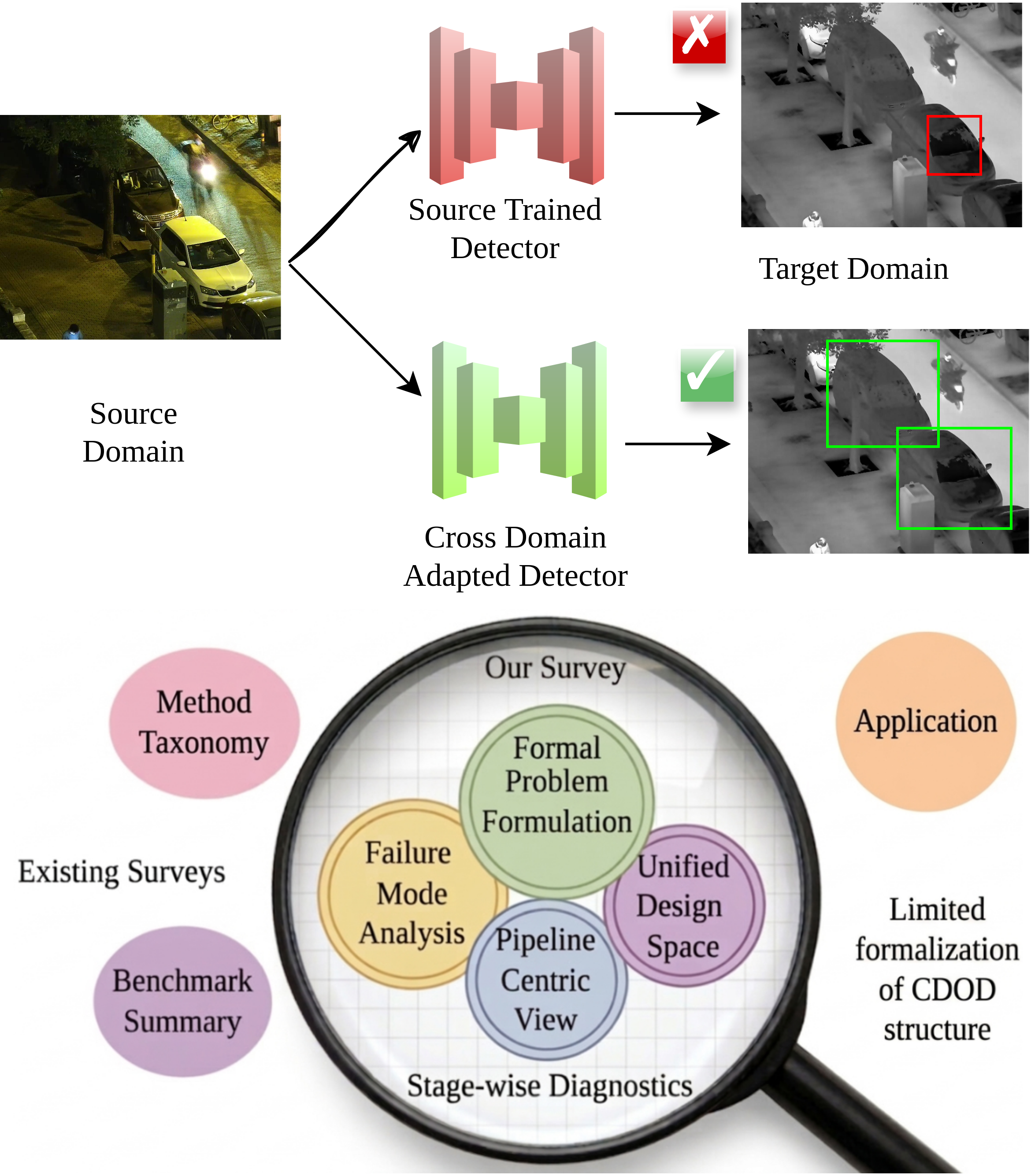}
\caption{Overview and motivation of this survey. Top: a source-trained detector degrades on the target domain under domain shift, while a cross-domain adapted detector improves target detections. Bottom: existing surveys mainly emphasize method taxonomy, benchmarks, and applications, whereas our survey focuses on four complementary pillars formal problem formulation, pipeline-centric analysis, failure-mode analysis, and a unified design space with stage-wise diagnostics.}
\label{fig:cross-domain-definition}
\end{figure}

Object detection models achieve impressive accuracy when trained and tested
under identical conditions, yet performance often degrades sharply in
deployment due to shifts in sensing conditions, weather, geography, or scene
composition \cite{wang2025cross,shi2025tdenet,liang2025perspective}. As shown
in Fig.~\ref{fig:cross-domain-definition}, cross-domain object detection
(CDOD) addresses this by adapting a source-trained model to a target domain
with a different distribution \cite{chen2018domain,zhu2019adapting}.
%%%
Unlike classification, detection solves two tightly coupled tasks
simultaneously: recognizing \emph{what} objects are present and localizing
\emph{where} they appear. Domain shift therefore reverberates through the
entire pipeline rather than striking a single point of failure, and preserving
semantic understanding does not automatically preserve geometric consistency
\cite{zhao2022task,zhang2022multiple}. This makes CDOD structurally harder
than classification-based adaptation.
%%%
A substantial body of work has explored adversarial feature alignment,
self-training, and domain generalization
\cite{chen2018domain,zhu2019adapting,saito2019strong,liu2024unbiased}, yet the
field remains fragmented. Methods are built on different assumptions and
evaluated on different benchmarks, so it is rarely clear which components of
the pipeline should be adapted or why a method succeeds in one setting but
fails in another.

% ---------------------------------------
\begin{table*}[t]
    \centering
    \caption{Comparison of representative surveys relevant to cross-domain object detection with respect to this work ($\checkmark$ = explicit coverage, $\triangle$ = partial coverage, $\times$ = no/unknown coverage).}
    \label{tab:survey-comparison}
    \scriptsize
    \setlength{\tabcolsep}{1.2pt}
\begin{tabularx}{\textwidth}{@{}>{\raggedright\arraybackslash}p{3cm}c@{\hspace{7pt}}c@{\hspace{7pt}}c@{\hspace{7pt}}c@{\hspace{7pt}}c@{}}
    \toprule
    \textbf{Survey} & \textbf{Year} & \textbf{Formal Problem} & \textbf{Pipeline} & \textbf{Failure} & \textbf{Unified Framework} \\
    \midrule
    Pan \& Yang~\cite{pan2009survey} & 2009 & $\checkmark$ & $\times$ & $\times$ & $\times$ \\
    Weiss et al.~\cite{weiss2016survey} & 2016 & $\checkmark$ & $\times$ & $\times$ & $\times$ \\
    Csurka~\cite{csurka2017domain} & 2017 & $\checkmark$ & $\times$ & $\times$ & $\times$ \\
    Li et al.~\cite{li2020deep} & 2020 & $\triangle$ & $\triangle$ & $\triangle$ & $\times$ \\
    Muzammul \& Li~\cite{muzammul2021survey} & 2021 & $\triangle$ & $\times$ & $\times$ & $\times$ \\
    Oza et al.~\cite{oza2023unsupervised} & 2023 & $\triangle$ & $\triangle$ & $\triangle$ & $\times$ \\
    Zou et al.~\cite{zou2023object} & 2023 & $\times$ & $\triangle$ & $\times$ & $\times$ \\
    Xu et al.~\cite{xu2025deep} & 2025 & $\triangle$ & $\times$ & $\triangle$ & $\times$ \\
    \midrule
    \textbf{Ours} & \textbf{2026} & \textbf{$\checkmark$} & \textbf{$\checkmark$} & \textbf{$\checkmark$} & \textbf{$\checkmark$} \\
    \bottomrule
    \end{tabularx}
    \end{table*}

\subsection{Contextualisation}

Object detection under domain shift is best understood as a connected pipeline,
not as an isolated feature-matching problem. A detector first extracts visual
features, then generates proposals, and finally classifies and refines them.
When features drift across domains, proposal quality also shifts, and the heads
must operate on weaker inputs \cite{chen2018domain,saito2019strong}. The
dependency is bidirectional: feature adaptation changes proposal statistics,
while poor proposals reduce the learning signal available to classification and
regression heads \cite{zhu2019adapting,saito2019strong}. In practice, robust
CDOD requires maintaining proposal coverage, feature discriminativity, and
stable prediction behavior together.

The shift itself is multi-causal. Covariate changes come from appearance
factors such as illumination, weather, sensor properties, or style
\cite{chen2018domain,zhu2019adapting}. Label-distribution shift appears when
class frequencies or category sets differ between source and target
\cite{zheng2025universal,pan2020exploring}. Feature misalignment weakens class
separation in learned representations \cite{vs2021mega,huang2022category}, and
contextual shift alters scale, layout, and background regularities
\cite{chen2021scale,wang2025sr}. These sources of shift reinforce each other;
for example, noisy pseudo-labels can bias representation updates, which then
produce even noisier pseudo-labels in later iterations
\cite{saito2019strong,chen2025refining}.

This is also why classification-style adaptation theory transfers only
partially to detection. Detection outputs are structured sets, and the data seen
by later stages depends on upstream model behavior rather than on a fixed input
distribution \cite{chen2018domain,zhao2022task}. Moreover, mAP is
non-decomposable, so apparent improvements in one component can hide failures in
another \cite{zhao2022task,zhang2022multiple}. A method taxonomy alone is
therefore not enough; a stage-aware, pipeline-centric view is needed to explain
when a CDOD method works and why it breaks.

\subsection{Contributions}

This survey organizes the field around the detection pipeline itself, analyzing
how domain shift propagates across stages and which invariants adaptation must
preserve. The main contributions are:
\begin{itemize}
    \item A formal formulation of CDOD as constrained, stage-coupled
    optimization over three invariants: proposal coverage, feature
    discriminativity, and calibration.
    \item A six-axis conceptual taxonomy that reveals systematic gaps in the
    current design space.
    \item A probabilistic pipeline decomposition explaining how shift propagates
    across stages and produces characteristic failure modes.
    \item A review of datasets, benchmarks, and evaluation protocols.
    \item An analysis of seven deep challenges and concrete future research
    directions.
\end{itemize}

Section~\ref{sec:overview} establishes the formal problem definition.
Section~\ref{sec:taxonomy} presents the taxonomy.
Sections~\ref{sec:synthesis} and~\ref{sec:discussion} synthesize insights and
discuss evaluation limitations. Sections~\ref{sec:datasets}
through~\ref{sec:failure-modes} cover datasets, challenges, and failure modes.
Section~\ref{sec:future-directions} outlines research directions and
Section~\ref{sec:conclusion} concludes.

% ---------------------------------------
\begin{figure*}[t]
\centering
\includegraphics[width=\textwidth]{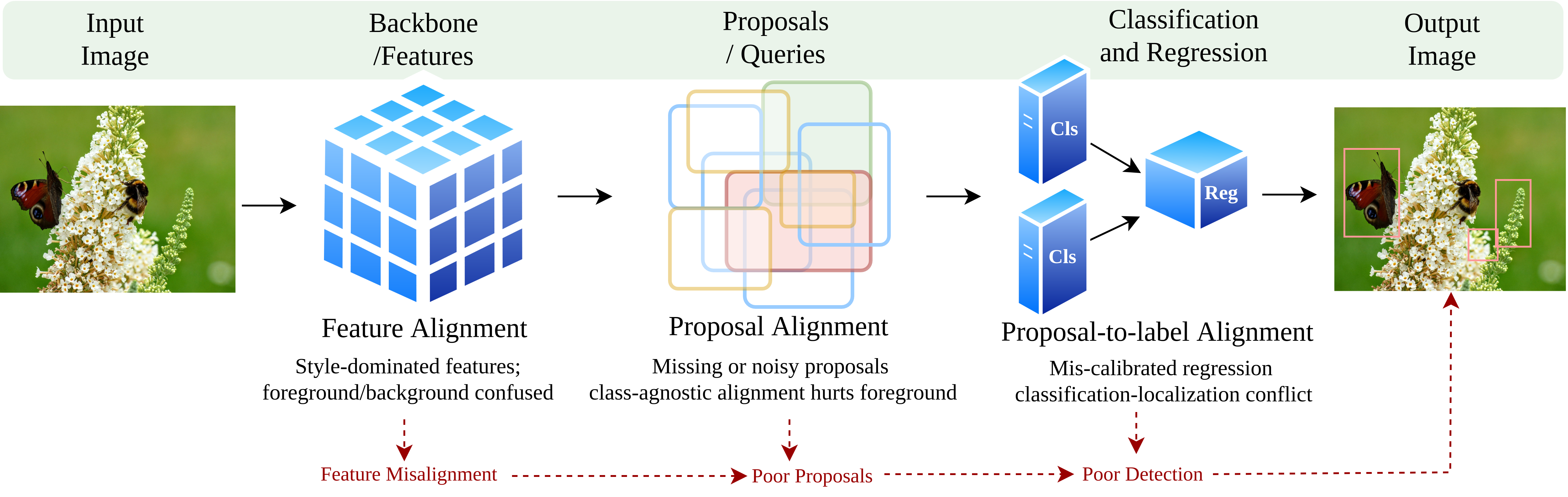}
\caption{Overview of the object detection pipeline highlighting three sources of misalignment: feature alignment (style-dominated features and foreground–background confusion), proposal alignment (missing or noisy proposals and class-agnostic mismatch), and proposal-to-label alignment (mis-calibrated regression and classification–localization conflict). These misalignments propagate through the pipeline, leading to poor proposals and ultimately degraded detection performance.}
\label{fig:design-space-compression}
\end{figure*}

\section{Overview}
\label{sec:overview}

This section sets up the common notation and the conceptual lens used throughout the survey. We focus on \emph{how} we will analyze methods: CDOD is treated as a stage-coupled problem, where feature extraction, proposal generation, and prediction heads interact under shift. The formal definition below gives precise symbols for domains, tasks, and detectors, so later sections can discuss what a method changes, what it assumes, and which part of the pipeline it improves or breaks.

\subsection{Formal Problem Definition}
\label{sec:formal-definition}
% ---------------------------------------

Following \cite{weiss2016survey,pan2009survey,csurka2017domain}, a domain $\mathcal{D}$ consists of a feature space $\mathcal{X}$ with marginal distribution $P(X)$, and a task $\mathcal{T}$ defined by label space $\mathcal{Y}$ and conditional distribution $P(Y|X)$. Given source domain $\mathcal{D}_s = \{\mathcal{X}_s, P_S(X)\}$ with task $\mathcal{T}_s = \{\mathcal{Y}_s, P_S(Y|X)\}$ and target domain $\mathcal{D}_t = \{\mathcal{X}_t, P_T(X)\}$ with task $\mathcal{T}_t = \{\mathcal{Y}_t, P_T(Y|X)\}$, when $\mathcal{D}_s \neq \mathcal{D}_t$ or $\mathcal{T}_s \neq \mathcal{T}_t$, knowledge transfer exploits related information from $\{\mathcal{D}_s, \mathcal{T}_s\}$ to learn $P_T(Y|X)$.

For cross-domain object detection, a detector $f$ maps an image $x \in \mathcal{X}$ to a set of detections:
\begin{equation}
f(x) = \{(b_i, c_i, s_i)\}_{i=1}^{N(x)},
\end{equation}
where $b_i$ is a bounding box, $c_i$ a class label, $s_i$ a confidence score, and $N(x)$ the number of detections. Let $P_S$ and $P_T$ denote source and target distributions over image-label pairs $(x,y)$. In the standard setting, we have labeled samples from $P_S$ and unlabeled (or partially labeled) samples from $P_T$ \cite{chen2018domain,saito2019strong}. Detectors produce proposals (explicit in two-stage detectors \cite{chen2018domain,saito2019strong}, implicit as anchors or queries in one-stage and query-based detectors), inducing a proposal distribution that depends on the input distribution. Under domain shift ($P_T \neq P_S$), the proposal distribution, feature distribution, and head behavior all change.

\emph{Cross-domain object detection} is the problem of minimizing expected detection loss on the target subject to maintaining three invariants:
\begin{equation}
\label{eq:grand-unifying}
\boxed{
\begin{aligned}
\min_f \quad & \mathbb{E}_{(x,y) \sim P_T}\bigl[L_{\text{cls}}(f(x), y) + L_{\text{reg}}(f(x), y)\bigr] \\
\text{s.t.} \quad & \text{Recall}_T(f) \geq \text{Recall}_S(f) - \epsilon \;\; \text{(proposal coverage)}, \\
& \mathrm{ECE}_T(f) \leq \delta \;\; \text{(calibration)}, \\
& \mathrm{Sep}_T(\phi(x)) \geq \gamma \;\; \text{(discriminativity)}.
\end{aligned}
}
\end{equation}
where $L_{\text{cls}}$ and $L_{\text{reg}}$ are classification and regression losses respectively, $\text{Recall}_T(f)$ and $\text{Recall}_S(f)$ are proposal recall on target and source (fraction of ground-truth objects with at least one proposal above an IoU threshold), $\mathrm{ECE}_T(f)$ is expected calibration error on target (measuring alignment between confidence scores and actual correctness), $\mathrm{Sep}_T(\phi(x))$ is a discriminativity measure on target features $\phi(x)$ for example, Fisher ratio or minimum inter-class margin, and $\epsilon$, $\delta$, $\gamma$ are tolerances set from source performance or application requirements \cite{chen2025gaussian,cai2024uncertainty}. In practice, $P_T$ is unknown and optimization uses unlabeled target images (and possibly a small set of labeled target samples) together with labeled source data \cite{nguyen2020domain,yao2025source,diamant2024confusing}.

This formulation makes explicit that domain adaptation for detection cannot succeed by aligning features alone \cite{zhu2019adapting,zhao2022task,zhang2022multiple}. It must preserve: (1) proposal coverage: proposals in the target domain must cover true objects with similar recall as in the source; (2) feature discriminativity: the representation must remain discriminative for foreground vs.\ background and class boundaries; and (3) regression calibration: the mapping from features to box deltas must remain geometrically consistent. Figure~\ref{fig:design-space-compression} provides a high-level overview of object detection pipeline highlighting feature, proposal, and proposal-to-label misalignment, and their impact on detection performance.

% ---------------------------------------
\subsection{Operationalizing Stage-Coupled CDOD}
\label{sec:operationalizing}
% ---------------------------------------

The discriminativity measure $\mathrm{Sep}_T(\phi(x))$ can be instantiated as: (a) \emph{Fisher ratio}: $\frac{\text{between-class variance}}{\text{within-class variance}}$ over proposal or patch features, with class from pseudo-labels or foreground/background; or (b) \emph{Margin}: minimum distance between class centroids in feature space, or minimum margin of a linear probe \cite{vs2021mega,jiang2025adaptive}. Both are measurable on target with pseudo-labels or foreground masks and drop when alignment blurs discriminativity \cite{zhu2019adapting,he2025differential}.

To evaluate which invariants are preserved or broken, stage-wise diagnostic metrics complement mAP \cite{zhao2022task,zhang2022multiple,he2025differential}: (1) \emph{Proposal stage}: $\mathcal{R}_{\mathrm{prop}}^T(\tau)$ - proposal recall at IoU $\tau$ on target; (2) \emph{Classification stage}: $\mathrm{Acc}_{\mathrm{cls}}^{\mathrm{oracle}}$ - classification accuracy given ground-truth boxes on target; (3) \emph{Regression stage}: $\bar{\mathcal{E}}_{\mathrm{loc}}^{\mathrm{TP}}(\tau)$ - mean localization error on matched true positives; (4) \emph{Calibration}: $\mathrm{ECE}_T$ - expected calibration error on target.  As shown in Fig.~\ref{fig:design-space-compression}, domain shift can affect multiple components of the detection pipeline, leading to compounded performance degradation.

% ---------------------------------------
\subsection{Pipeline as a Dependency Graph: Probabilistic Decomposition}
\label{sec:pipeline-dependency}
% ---------------------------------------
\emph{This subsection is the hinge of the survey.} Failure propagation across stages can be stated in one equation. The target detection distribution factors as:
\begin{equation}
\label{eq:decomposition}
P_T(y \mid x) = \int P_T(y \mid b, x)\, P_T(b \mid x)\, \mathrm{d}b.
\end{equation}
Here $y$ is the set of outputs (box, class), $b$ indexes the set of proposals, $P_T(b \mid x)$ is the proposal distribution given the image, and $P_T(y \mid b, x)$ is the conditional distribution modeled by the detection head(classification and regression given proposals and image). Two observations sharpen the implications.

\emph{Observation 1.} If proposal recall degrades under shift (i.e., $P_T(b \mid x)$ places little mass on correct object locations), then \emph{no} adaptation restricted to improving $P_T(y \mid b, x)$ alone can recover target risk \cite{saito2019strong,li2023learning}. The head receives a biased or impoverished set of proposals; optimizing the head conditional cannot create proposals that were never generated. The ceiling is structural.

\emph{Observation 2.} Feature alignment that alters $P_T(b \mid x)$, by changing backbone or proposal inputs, changes the effective input distribution to the head \cite{zhang2024pseudo,diamant2024confusing}. Any method that assumes the head can be adapted in isolation for example, head-only fine-tuning or head alignment, implicitly assumes a fixed or transferable $P_T(b \mid x)$. When alignment shifts that distribution, head-only adaptation is mis-specified--the assumptions of head-only adaptation are violated \cite{saito2019strong,xu2022h2fa}. In particular, most CDOD methods implicitly assume $P_T(b \mid x) \approx P_S(b \mid x)$; when this assumption fails, head-level alignment cannot compensate for proposal-level shift.

Thus: adaptation objectives are \emph{coupled}; improving one stage changes the distribution the next stage sees \cite{saito2019strong,yang2025versatile,zhao2022task}. Eq.~\ref{eq:decomposition} elevates the pipeline view from diagram to a probabilistic argument (Observation 1 and 2).

% ---------------------------------------
\subsection{Structural Causes of Domain Shift}
% ---------------------------------------

Domain shift in object detection is not a single phenomenon. It decomposes into several structural causes, each with distinct implications for where and how to adapt.

\emph{Covariate shift} refers to a change in the marginal distribution of inputs $P(X)$ while the conditional $P(Y \mid X)$ is assumed stable. In detection, this manifests as changes in image style, resolution, illumination, or sensor characteristics for example, optical vs.\ synthetic aperture radar, clear vs.\ foggy weather. Most feature-alignment and style-transfer methods target covariate shift explicitly \cite{chen2018domain,zhu2019adapting,zheng2020cross,xu2020cross,li2022scan,chen2021scale,deng2021unbiased,nguyen2020domain,wang2021afan,do2022exploiting,song2024cross,piao2023unsupervised,kay2024align}.

\emph{Label (and concept) distribution shift} refers to changes in $P(Y)$ or in the set of classes present across domains. In closed-set CDOD, class proportions may differ for example, more pedestrians in one city than another \cite{chen2018domain,zheng2020cross}; in open-set or partial-set settings, the target may contain classes absent in the source or vice versa \cite{zheng2025universal,pan2020exploring}. Methods that perform category-agnostic alignment can suffer negative transfer when label distributions differ; universal and open-set DAOD explicitly distinguish between shared vs.\ private categories \cite{zheng2025universal,pan2020exploring}.

\emph{Feature misalignment} is the failure of the learned representation to remain discriminative or geometrically consistent across domains. Even when covariate and label shift are addressed, the internal feature space can become distorted: style may dominate semantics, or foreground and background may be poorly separated \cite{zhu2019adapting,vs2021mega,huang2022category,cai2024uncertainty,wu2021vector}. This is a consequence of \emph{where} alignment is applied for example, image-level vs.\ instance-level and \emph{what} objective is used for example, domain confusion vs.\ task-specific consistency.

\emph{Contextual shift} captures changes in scene layout, object scale, density, occlusion patterns, and background semantics \cite{chen2021scale,wang2025sr,liang2025perspective}. The same object class may appear at different scales or in different surroundings; the statistical relationship between context and object can change. Detection is inherently context-dependent for example, ``car'' in a street vs.\ in a parking lot \cite{zhang2019category,iqbal2021leveraging}, so contextual shift directly affects both proposal generation and classification \cite{chen2021scale,wang2025sr}.

\emph{Annotation bias} arises when source labels are incomplete, noisy, or defined under different protocols for example, different box tightness, different class granularity. Pseudo-labels and self-training inherit and can amplify these biases in the target domain \cite{saito2019strong,yang2025versatile,chen2025refining,wei2025multi,kim2024vlm}. This cause is often overlooked in formal definitions of domain shift but is central to the reliability of adaptation methods that rely on source-trained models or generated target labels.

These causes are not independent: covariate and contextual shift jointly affect feature quality \cite{wang2025sr,chen2021scale}; label distribution shift interacts with annotation bias in self-training \cite{saito2019strong,zheng2025universal}. A complete view of CDOD must account for all five \cite{luo2025mas,wang2025multidimensional}.

% ---------------------------------------
\subsection{Why Object Detection Is Harder Than Classification in Domain Transfer}
% ---------------------------------------

Domain adaptation for object detection is fundamentally harder than domain adaptation for image classification \cite{chen2018domain,zhao2022task}. First, detection has two coupled outputs--classification and localization--that share representation but respond differently to shift. Classification may degrade due to semantic or style confusion; regression degrades when feature scale, object scale, or context changes. Aligning ``features'' for classification can leave regression poorly calibrated, and vice versa; classification-focused DA does not face this dual-output structure \cite{zhao2022task,zhang2022multiple}. Second, the proposal stage creates a bottleneck: in two-stage detectors, the region proposal network (RPN) or equivalent must produce reliable candidate boxes in the target domain, and if proposals are missing or biased, downstream alignment cannot recover \cite{chen2018domain,saito2019strong}. In one-stage and query-based detectors, the analogue is the set of anchors or queries that effectively ``propose'' regions \cite{lavoie2025large,yang2025fsda}. No such intermediate structure exists in classification. Third, foreground and background are asymmetric: classification assumes a single object (or a fixed grid of patches), whereas detection must separate foreground from background at every location. Domain shift can alter the appearance of both; global or image-level alignment often equalizes foreground and background, harming recall and localization \cite{zhu2019adapting,vs2021mega,wu2021vector}. Fourth, spatial and geometric consistency matter detection requires that the same object yield consistent boxes across domains, and regression heads are sensitive to the distribution of features they receive in a way that standard classification is not \cite{zhang2020multi,zhou2024dual}. Finally, annotation cost and protocol vary more severely for detection; cross-domain detection must contend with no target labels (UDA), few labels (SSDA), or source-free settings \cite{yao2025source,diamant2024confusing,yu2019unsupervised,yang2025fsda,zhao2025fsdaod,shangguan2025cross}, all under the added complexity of spatial annotations.

% ---------------------------------------

\subsection{Why Classification DA Bounds Do Not Carry Over to Detection}
\label{sec:why-classification-bounds-break}
% ---------------------------------------
Classification DA theory by Ben-David et al. \cite{ben2010theory} bounds target error by source error plus a domain divergence plus ideal joint error. Detection adaptation does not directly inherit these bounds \cite{chen2018domain,zhao2022task} for four structural reasons. (1) \emph{Output is a structured set}: detections are sets of (box, class, score); matching and loss (mAP) depend on assignment and ranking, not a single label per example \cite{zhang2021c2fda,liu2024object}. (2) \emph{Loss is non-decomposable}: As a ranking-based metric (area under the precision–recall curve), mAP does not admit a decomposition into independent per-sample terms, since it is not linear in per-example errors\cite{zhao2022task}. (3) \emph{Proposal distribution is endogenous}: the input to the classification/regression heads is $(b, \phi(x))$ where $b$ is produced by the same model; the effective distribution on which the head operates shifts when we change the model (Sec.~\ref{sec:pipeline-dependency}). Ben-David-style bounds assume fixed input distributions. (4) \emph{Conditional risk depends on proposal recall}: even if one could write a bound, the target risk of the head is conditioned on the proposal distribution; when recall drops, the head sees a biased sample and the bound would need to account for that coupling. Thus classification DA theory does not directly apply \cite{chen2018domain,he2024recalling}; new bounds for detection (ranking-based metrics, endogenous proposals, structured outputs) are needed.

% ---------------------------------------

\subsection{Fragmentation and the Need for a Unifying Lens}
% ---------------------------------------

The CDOD literature has fragmented into camps (feature alignment, pseudo-label/self-training, localization-centric) that rarely compose and seldom ask which stage is the bottleneck \cite{zhu2019adapting,saito2019strong,zhao2022task}. A deeper tension is that objectives improving domain invariance can conflict with those preserving detection performance \cite{zhu2019adapting,he2025differential}. Existing surveys organize by technique or setting and do not offer a formal problem definition or pipeline-centric view \cite{chen2018domain,zhao2022task}, so ``solving'' CDOD remains ill-defined. Table~\ref{tab:survey-comparison} shows this gap and positions our survey as a unified formulation with pipeline-centric and invariant-aware analysis.

\subsection{Conceptual Lens and Pipeline View}
% ---------------------------------------

We recast CDOD as constrained, stage-coupled optimization (Sec.~\ref{sec:formal-definition}, Eq.~\ref{eq:grand-unifying}); the pipeline decomposition (Eq.~\ref{eq:decomposition}, Sec.~\ref{sec:pipeline-dependency}) is the central structural result. We organize existing work by which stage they target and which shift type they assume, synthesize failure modes, and propose stage-wise diagnostics and research directions. As shown in Table~\ref{tab:survey-comparison}, this survey uniquely provides a formal constrained optimization framework, probabilistic pipeline decomposition, design-space compression analysis, and explicit treatment of invariant preservation--distinguishing it from prior surveys that focus primarily on method categorization or application-specific analysis.

% =============================================================================
% =============================================================================

\section{A Conceptual Taxonomy for Cross-Domain Object Detection}
\label{sec:taxonomy}

Existing surveys categorize by implementation (feature vs.\ pixel, adversarial vs.\ self-training) \cite{inoue2018cross,liu2019improving}. That obscures the philosophical and structural choices that define a method. We propose six axes; placing a method along them reveals assumptions, scope, and where theory is lacking. This taxonomy is used as an analytic tool to identify what each method changes, what assumptions it makes, and which pipeline stage it primarily targets. Using the taxonomy, we observe that most of the literature clusters in one region--design-space compression is visible.

% ---------------------------------------
\subsection{Alignment vs.\ Invariance vs.\ Robustness-Based Paradigms}
\label{sec:taxonomy:paradigms}
% ---------------------------------------

\begin{figure}[t]
\centering
\includegraphics[width=\columnwidth]{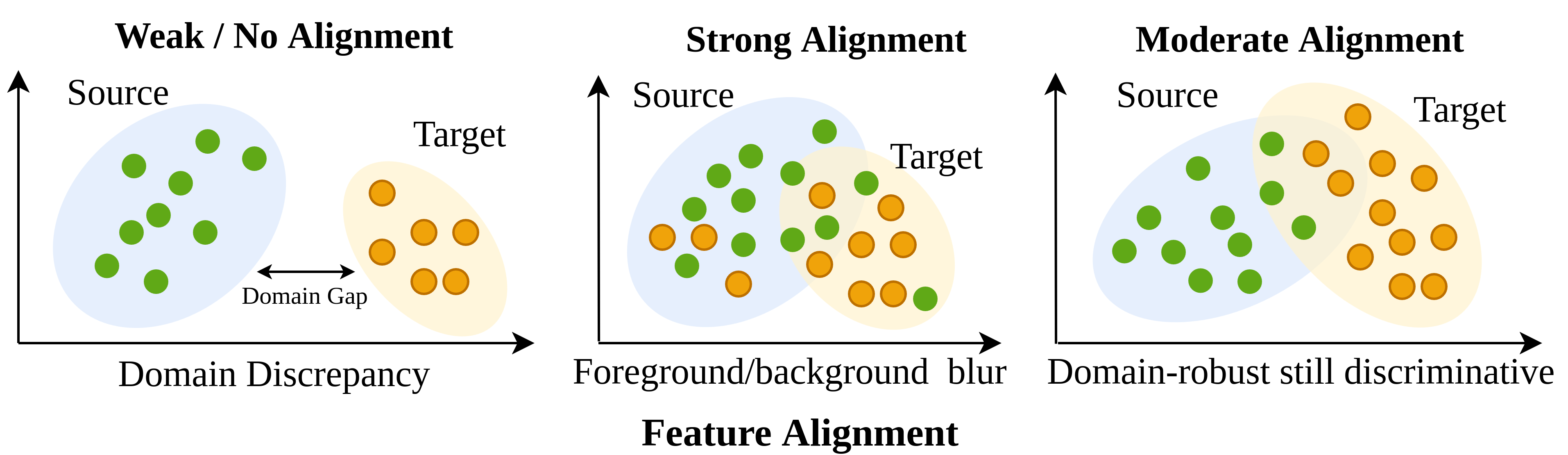}
\caption{The alignment-discriminativity tension: strong alignment can erase the feature structure that detection needs, so objectives must balance alignment with task-relevant discriminativity.}
\label{fig:alignment-discriminativity}
\end{figure}

\emph{Alignment-based} methods modify source/target representations to reduce distribution mismatch (pixel, feature, or latent-space alignment) \cite{chen2018domain,zhu2019adapting,zheng2020cross}. Their core assumption is that once domain discrepancy is reduced, the same detector can transfer with limited target-specific redesign. In practice, they primarily target the feature stage (and sometimes proposal inputs), with indirect effects on heads. This can improve transfer when source and target are related, but strong alignment can erase task-relevant structure and hurt localization \cite{zhu2019adapting,he2025differential}. As illustrated in Fig.~\ref{fig:alignment-discriminativity}, weak alignment leaves a large domain gap, overly strong alignment blurs discriminative structure, and moderate alignment offers a better trade-off.

\emph{Invariance-based} methods modify representation learning or decision boundaries to keep task-relevant factors stable across domains \cite{liu2019improving,biswas2024domain}. Their assumption is that the chosen invariants are truly domain-stable and sufficient for detection. They mostly target feature semantics and classifier behavior, but can underperform when the selected invariances do not match real deployment shifts \cite{biswas2024domain}.

\emph{Robustness-based} methods modify training data or objectives so the detector remains reliable over a family of plausible shifts (augmentation, style diversification, DG) without target adaptation data \cite{liu2024unbiased,geng2026cen,saoud2023mars,saoud2024real}. Their assumption is that training-time domain diversity approximates test-time conditions. They affect the full pipeline through robust feature learning and more stable proposal/head behavior, but often trade some in-domain accuracy for better out-of-domain stability \cite{liu2024unbiased,geng2026cen}.

Viewed through this lens, the three paradigms differ not only by technique but by intervention point, assumption set, and failure mode. Bounds linking alignment/invariance/robustness quality to detection risk remain limited, and paradigm composition for example, align style while preserving invariant content is still underexplored \cite{tulu2025wct,xu2024dst}. Fig.~\ref{fig:alignment-discriminativity} illustrates the alignment-discriminativity tension.

% ---------------------------------------
\subsection{Geometry vs.\ Semantic-Preserving Adaptation}
\label{sec:taxonomy:geometry-semantics}
% ---------------------------------------

\emph{Geometry-preserving} methods mainly change how localization signals are transferred (box coordinates, scales, aspect ratios, proposal geometry) \cite{cheng2022anchor,zhou2025ccanet,niu2023object}. They assume geometric relations remain transferable across domains, and they primarily target the proposal and regression stages. Their risk is that preserving geometry alone can leave class boundaries under-adapted when semantic shift is strong.

\emph{Semantic-preserving} methods mainly change feature/class representations so category meaning stays stable across domains \cite{zhao2022task,zhang2022multiple}. They assume semantic structure is the main bottleneck and mostly target feature and classification stages. Their risk is regression mis-calibration when localization-specific shift is not addressed.

Under this axis, the key analytical point is stage imbalance: geometry-only strategies can miss semantic adaptation, while semantic-only strategies can miss localization stability. Robust CDOD requires both, but explicit cls/reg balancing is still uncommon \cite{cheng2022anchor,zhou2025ccanet}.

\begin{figure}[t]
\centering
\includegraphics[width=\columnwidth]{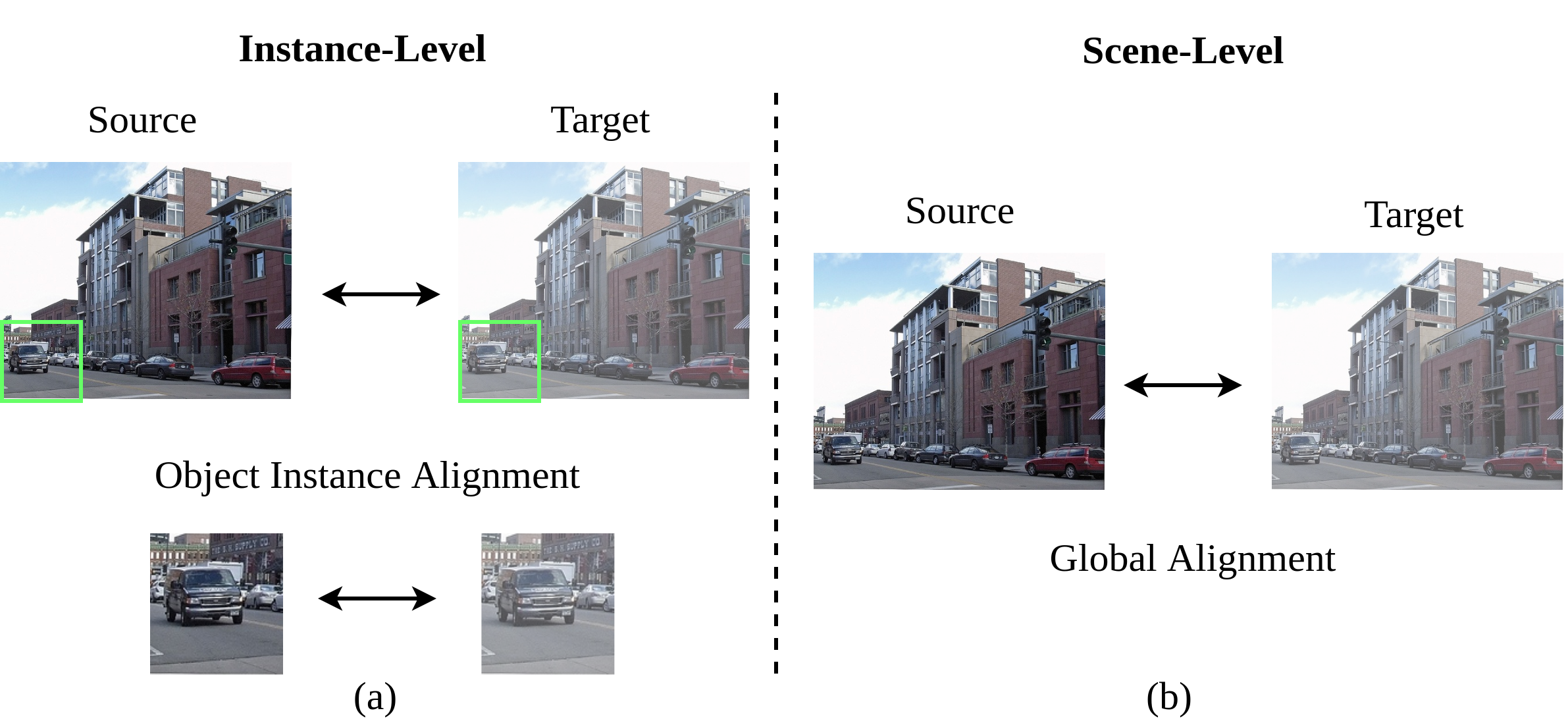}
\caption{Instance-level versus scene-level adaptation: adapting localized instances versus global scene context.}
\label{fig:instance-scene-adaptation}
\end{figure}

% ---------------------------------------
\subsection{Implicit vs.\ Explicit Distribution Modeling}
\label{sec:taxonomy:implicit-explicit}
% ---------------------------------------

\emph{Implicit} methods change optimization objectives (domain confusion, consistency, contrastive losses) without directly estimating source/target densities \cite{zhu2019adapting,zheng2020cross,saito2019strong,tulu2025wct}. They assume these objectives are sufficient proxies for transfer and usually target feature alignment and pseudo-label refinement. Their main risk is objective-driven collapse or over-smoothing when confusion is achieved without preserving detection structure.

\emph{Explicit} methods change the adaptation process by modeling distributions or statistics directly (prototypes, Gaussians, generative components) \cite{jiang2025adaptive,xu2020cross,chen2025gaussian}. They assume the chosen model class is adequate for real shifts and typically target feature calibration and label selection quality. Their main risk is misspecification and extra computational overhead.

Analytically, this axis separates \emph{how evidence is represented}: implicit methods are scalable but opaque, explicit methods are interpretable but brittle when modeling assumptions fail. Hybrid designs remain promising but are still mostly heuristic \cite{kennerley2025bridging}.

% ---------------------------------------
\subsection{Instance vs.\ Scene-Level Adaptation}
\label{sec:taxonomy:instance-scene}
% ---------------------------------------

\emph{Instance-level} methods change localized object representations (RoIs, proposal features, object-centric crops) \cite{zhu2019adapting,jiao2022dual,do2022exploiting}. As shown in Fig.~\ref{fig:instance-scene-adaptation}(a), they assume proposals in the target domain are sufficiently reliable and primarily target proposal/head interactions. Their strength is foreground focus; their failure mode is proposal bias propagation when target proposals are poor.

\emph{Scene-level} methods change global image features or style statistics \cite{chen2018domain,zheng2020cross}. As shown in Fig.~\ref{fig:instance-scene-adaptation}(b), they assume global context alignment is enough to improve downstream detection and primarily target the backbone feature stage. Their failure mode is over-aligning background and foreground together, which can reduce discriminativity for small or rare objects.

This axis clarifies adaptation granularity as a design choice: local methods are precise but proposal-dependent, global methods are stable but coarse. The coupling between scene-level alignment and proposal quality remains under-characterized \cite{gao2022progressive,liu2025don}.

% ---------------------------------------
\subsection{Closed vs.\ Open-Set vs.\ Universal Domain Shift}
\label{sec:taxonomy:closed-open-universal}
% ---------------------------------------

\emph{Closed-set} methods change adaptation objectives under the assumption of identical source/target label spaces; they mainly target covariate transfer and usually optimize feature/head alignment \cite{chen2018domain,zheng2020cross,saito2019strong}. Their failure mode is negative transfer when unseen target classes are forced into known source categories.

\emph{Open-set} methods change the objective by adding unknown-class handling (rejection, thresholding, separation of shared vs.\ private classes). They assume unknowns can be separated without target labels and primarily target classification calibration and decision boundaries \cite{zheng2025universal,pan2020exploring}. Their failure mode is threshold sensitivity and unstable unknown detection.

\emph{Universal} methods further change the problem setup to handle closed-, open-, and partial-set regimes together, often without prior regime knowledge \cite{zheng2025universal}. They assume robust shared/private discovery is feasible under weak supervision and target both representation and decision stages. Their failure mode is compounding uncertainty from clustering, pseudo-labeling, and class-partition estimation.

\begin{figure}[t]
\centering
\includegraphics[width=\columnwidth]{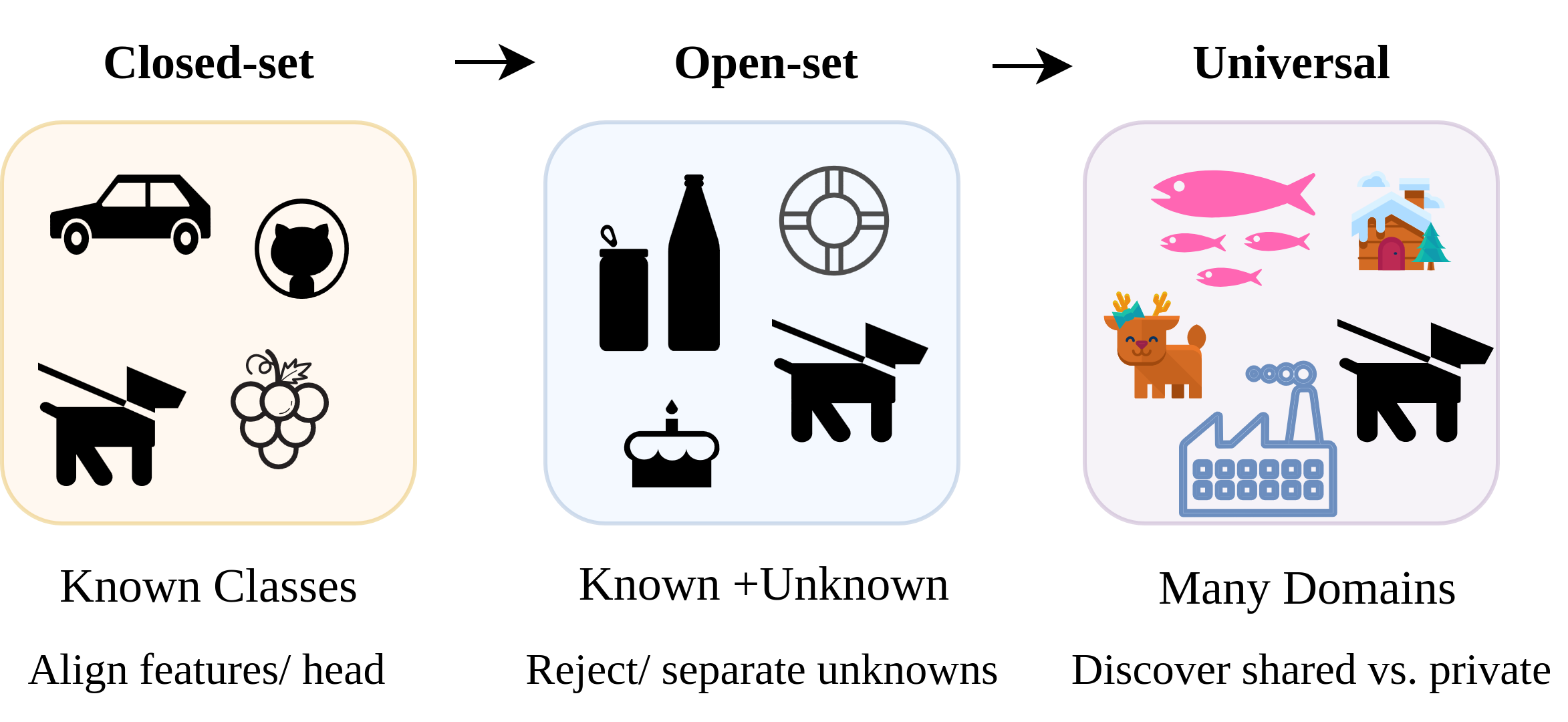}
\caption{Progression from closed-set to open-set to universal domain shift: closed-set assumes identical class sets; open-set allows target-private classes; universal handles all scenarios including partial-set.}
\label{fig:closed-open-universal}
\end{figure}

Figure~\ref{fig:closed-open-universal} visualizes this progression in assumptions and difficulty. Analytically, this axis highlights that label-space assumptions are first-order design choices, not minor implementation details.

% ---------------------------------------
\subsection{Causal vs.\ Correlational Adaptation}
\label{sec:taxonomy:causal-correlational}
% ---------------------------------------

\emph{Correlational} methods change representations or decision functions by matching observed statistics across domains (differences in $P(X)$, $P(Y)$, or $P(X,Y)$). They assume statistical similarity implies transferability and mainly target feature alignment/classification behavior. Their failure mode is spurious alignment: performance can drop when new shifts break learned correlations.

\emph{Causal} methods change the modeling perspective by introducing structural assumptions about the causes of shift for example, style, sensor, context and aiming for invariance under interventions \cite{zhang2022multiple,kennerley2025bridging}. They assume causal factors can be identified well enough for robust adaptation and potentially affect all pipeline stages. Their failure mode is model misspecification: wrong causal structure can degrade performance more than purely correlational approaches.

Under this axis, the core analytical distinction is not method family but \emph{assumption depth}: correlational methods optimize observed associations, while causal methods require explicit structural commitments that are still rare and difficult to validate in detection \cite{kennerley2025bridging}.

% ---------------------------------------
\subsection{Why Most Methods Occupy the Same Region of the Taxonomy}
\label{sec:taxonomy:compression}
% ---------------------------------------

\emph{Clustering and design-space compression.} The six axes, show that \emph{most CDOD methods occupy the same region}: alignment-based, implicit, closed-set, correlational, instance- or scene-level. Adversarial DA \cite{zhu2019adapting,he2025differential,cheng2025wmfa}, self-training \cite{saito2019strong,wang2025unsupervised}, teacher-student \cite{yang2025versatile,zhao2024taming}, and most ``SOTA'' work fall here. Invariance and robustness \cite{liu2024unbiased,geng2026cen} are minorities; explicit modeling \cite{jiang2025adaptive,xu2020cross} is a fraction; open-set and universal \cite{zheng2025universal} are niche; causal is almost absent. \emph{This is design-space compression.} Method development has concentrated heavily in a narrow region of the design space, with repeated refinements within this cluster. That corner fits the benchmark regime (one source, one target, unlabeled target, synthetic-to-real, mAP) \cite{chen2018domain,zheng2020cross,saito2019strong}. Consequence: incremental progress within a narrow design space \cite{vs2021mega,he2025differential}; entire regions (explicit, open-set, universal, causal) underexplored \cite{jiang2025adaptive,zheng2025universal,xu2024dst}. The taxonomy forces the question: why so few methods outside this cluster, and what would it take to populate the rest?

The axes are not independent: alignment-based methods are usually correlational and closed-set. An adversarial feature-aligner \cite{zhu2019adapting,he2025differential} therefore sits in the dominant cluster, while style-transfer DG \cite{liu2024unbiased,tulu2025wct} at least departs from the alignment/robustness axis concentration. This cross-axis coupling further explains why many methods fail under stronger shifts and why large parts of the design space remain weakly explored.

% =============================================================================
% ---------------------------------------

\section{Synthesis and Key Insights}
\label{sec:synthesis}

Table~\ref{tab:taxonomy-comparison} and Table~\ref{tab:diagnostic} summarize how representative methods populate the CDOD taxonomy and identify which invariants are preserved or ignored across approaches. The synthesis below distills the main insights.

% ---------------------------------------
\subsection{Key Insights}
% ---------------------------------------

\emph{CDOD is a pipeline-level problem, not a feature-level problem.} Feature-only alignment assumes that proposals and detection heads transfer across domains; when this assumption fails, performance gains become inconsistent\cite{chen2018domain,zhu2019adapting,saito2019strong,zhao2022task}.

\emph{Domain-invariant features are not sufficient for domain-robust detection.} The objective should be maintaining \emph{what} and \emph{where} under shift; alignment must preserve task-relevant structure and evaluation must separate classification and localization \cite{zhu2019adapting,he2025differential,zhao2022task}.

\emph{mAP as sole metric confounds stages} (Sec.~\ref{sec:discussion}); taxonomy by structural choices exposes design-space compression (Sec.~\ref{sec:taxonomy:compression}, Table~\ref{tab:taxonomy-comparison}); failure modes and hidden assumptions should be explicit so practitioners can anticipate when a method fails \cite{biswas2024domain,li2024domain}.

% ---------------------------------------
\subsection{Empirical Observations and Insights}
% ---------------------------------------

The following insights are grounded in Eq.~\ref{eq:grand-unifying}, Eq.~\ref{eq:decomposition}, and the taxonomy, and they clarify current boundaries of evidence.

\emph{Most feature-alignment methods improve classification at the expense of localization, but this trade-off is rarely measured or reported.}
Strong domain alignment for example, global adversarial training \cite{zhu2019adapting,he2025differential} can achieve domain confusion by smoothing or compressing the feature space in ways that harm regression sensitivity. Because benchmarks report only mAP, a method that improves classification and slightly hurts localization may still ``win''; the regression degradation is hidden. Disentangled metrics would likely show that many SOTA methods are classification-centric and leave localization under-adapted.

\emph{Self-training and pseudo-labeling in CDOD are under-theorized and prone to confirmation bias; the field over-relies on them for lack of better target-side signals.}
Self-training provides a target-side supervisory signal when no labels exist, but it can lock onto wrong predictions and amplify source bias \cite{saito2019strong,chen2025refining,wei2025multi}. Theoretical guarantees for example, under what conditions pseudo-labels converge to true labels are scarce for detection. The success of self-training \cite{yang2025versatile,zhao2024taming}may reflect the absence of alternatives rather than its intrinsic suitability; investment in alternative target-side signals for example, foundation-model guidance \cite{vcr2025foundation}, contrastive objectives could yield more robust adaptation.

\emph{Domain generalization for detection is under-investigated relative to UDA; the community has over-indexed on the ``target data available at adaptation time'' setting.}
Unsupervised domain adaptation assumes unlabeled target data at adaptation time \cite{chen2018domain,saito2019strong}. In many real scenarios (new sensor, new geography, new deployment), target data may be scarce or unavailable until after deployment \cite{saoud2023mars,liu2024unbiased}. Domain generalization (no target data) is harder but more broadly applicable. The relative effort spent on UDA vs.\ DG \cite{liu2024unbiased,geng2026cen} does not match the relative need; DG for detection deserves more attention and more realistic benchmarks.

\emph{Causal and correlational adaptation are not yet meaningfully distinguished in practice; most ``causal'' CDOD work is still correlational with a causal narrative.}
Causal domain adaptation posits interventions on causes of shift for example, style to preserve causal effects for example, content $\rightarrow$ label \cite{zhang2022multiple}. In practice, most CDOD methods match statistics or learn invariances without a formal causal model or intervention \cite{tulu2025wct,kennerley2025bridging}. Claiming that ``we separate style and content'' is not the same as specifying a causal graph and estimating causal effects. Rigorous causal CDOD would require explicit graphs, identifiability analysis, and intervention-based evaluation; until then, the causal vs.\ correlational axis is largely conceptual.

\emph{Greater emphasis on diagnostic and compositional studies may yield higher long-term returns than incremental method variants.}
The rate of new CDOD methods outstrips the rate of diagnostic work that explains \emph{why} a method works or fails for example, which stage improved, which shift type was addressed, which assumption was violated \cite{zhao2022task,zhang2022multiple,he2025differential}. While many works include standard ablations, rigorous diagnostic studies for example, stage-wise isolation, oracle analyses, or controlled evaluation of shift types remain limited. A shift toward diagnostic and compositional research would improve interpretability and composability and reduce redundant point solutions.

\FloatBarrier

\begin{table*}[t]
    \centering
    \caption{Taxonomy coding of representative CDOD methods. Columns and values: Stage (feature/head/regression), Set = Setting (UDA: unsupervised domain adaptation, DG: domain generalization, s-free: source-free, few-shot, univ: universal), Par = Paradigm ($\leftrightarrow$: alignment, $\mathcal{R}$: robustness, $\phi$: hybrid), Rep = Representation ($\triangle$: semantic, $\diamond$: geometric), Mod = Modeling ($\circ$: implicit, $\bullet$: explicit), Lbl = Label Space ($\blacksquare$: closed-set, $\square$: open-set, $\phi$: hybrid), Gran = Granularity ($\circ$: instance-level, $\Box$: scene-level, $\phi$: mixed/uncertain), and Rsn = Reasoning ($\sim$: correlational, $\rightarrow$: causal, $\phi$: mixed/uncertain).}
    \label{tab:taxonomy-comparison}
    \scriptsize
    % Tighten horizontal padding between columns a little.
    \setlength{\tabcolsep}{1.2pt}
    \renewcommand{\arraystretch}{0.95}
\begin{tabularx}{\linewidth}{@{}>{\raggedright\arraybackslash}p{3.2cm}ccccccccc@{}}
    \toprule
    \textbf{Method} & \textbf{Year} & \textbf{Stage} & \textbf{Set} & \textbf{Par} & \textbf{Rep} & \textbf{Mod} & \textbf{Lbl} & \textbf{Gran} & \textbf{Rsn} \\
    \midrule
    DA-Faster~\cite{chen2018domain} & 2018 & feature & UDA & $\leftrightarrow$ & $\triangle$ & $\circ$ & $\blacksquare$ & $\phi$ & $\sim$ \\
    Strong-Weak~\cite{saito2019strong} & 2019 & feat.+head & UDA & $\leftrightarrow$ & $\triangle$ & $\circ$ & $\blacksquare$ & $\phi$ & $\sim$ \\
    Yu et al.~\cite{yu2019unsupervised} & 2019 & feat.+head & UDA & $\leftrightarrow$ & $\triangle$ & $\circ$ & $\blacksquare$ & $\phi$ & $\sim$ \\
    Category anchor~\cite{zhang2019category} & 2019 & feature & UDA & $\leftrightarrow$ & $\diamond$ & $\circ$ & $\blacksquare$ & $\phi$ & $\sim$ \\
    Selective~\cite{zhu2019adapting} & 2019 & feature & UDA & $\leftrightarrow$ & $\triangle$ & $\circ$ & $\blacksquare$ & $\circ$ & $\sim$ \\
    Nguyen et al.~\cite{nguyen2020domain} & 2020 & feature & UDA & $\leftrightarrow$ & $\diamond$ & $\circ$ & $\blacksquare$ & $\circ$ & $\sim$ \\
    Xu et al.~\cite{xu2020cross} & 2020 & feature & UDA & $\leftrightarrow$ & $\triangle$ & $\bullet$ & $\blacksquare$ & $\phi$ & $\sim$ \\
    Zhang et al.~\cite{zhang2020multi} & 2020 & feature & UDA & $\leftrightarrow$ & $\diamond$ & $\circ$ & $\blacksquare$ & $\circ$ & $\sim$ \\
    CR-DA~\cite{zheng2020cross} & 2020 & feature & UDA & $\leftrightarrow$ & $\triangle$ & $\circ$ & $\blacksquare$ & $\phi$ & $\sim$ \\
    Scale DA~\cite{chen2021scale} & 2021 & feature & UDA & $\leftrightarrow$ & $\diamond$ & $\circ$ & $\blacksquare$ & $\Box$ & $\sim$ \\
    Unbiased MT~\cite{deng2021unbiased} & 2021 & feat.+head & UDA & $\leftrightarrow$ & $\triangle$ & $\circ$ & $\blacksquare$ & $\phi$ & $\sim$ \\
    Mega-CDA~\cite{vs2021mega} & 2021 & feature & UDA & $\leftrightarrow$ & $\triangle$ & $\bullet$ & $\blacksquare$ & $\phi$ & $\sim$ \\
    AFAN~\cite{wang2021afan} & 2021 & feature & UDA & $\leftrightarrow$ & $\triangle$ & $\circ$ & $\blacksquare$ & $\Box$ & $\sim$ \\
    C2FDA~\cite{zhang2021c2fda} & 2021 & feature & UDA & $\leftrightarrow$ & $\diamond$ & $\circ$ & $\blacksquare$ & $\phi$ & $\sim$ \\
    Cheng et al.~\cite{cheng2022anchor} & 2022 & feature & UDA & $\leftrightarrow$ & $\diamond$ & $\circ$ & $\blacksquare$ & $\circ$ & $\sim$ \\
    Do et al.~\cite{do2022exploiting} & 2022 & feature & UDA & $\leftrightarrow$ & $\triangle$ & $\circ$ & $\blacksquare$ & $\Box$ & $\sim$ \\
    Progressive~\cite{gao2022progressive} & 2022 & feature & UDA & $\leftrightarrow$ & $\triangle$ & $\circ$ & $\blacksquare$ & $\Box$ & $\sim$ \\
    Category contrast~\cite{huang2022category} & 2022 & feature & UDA & $\leftrightarrow$ & $\triangle$ & $\circ$ & $\blacksquare$ & $\phi$ & $\sim$ \\
    Dual inst.~\cite{jiao2022dual} & 2022 & feat.+head & UDA & $\leftrightarrow$ & $\triangle$ & $\circ$ & $\blacksquare$ & $\circ$ & $\sim$ \\
    S-DAYOLO~\cite{li2022cross} & 2022 & feature & UDA & $\leftrightarrow$ & $\triangle$ & $\circ$ & $\blacksquare$ & $\phi$ & $\sim$ \\
    SCAN~\cite{li2022scan} & 2022 & feature & UDA & $\leftrightarrow$ & $\triangle$ & $\circ$ & $\blacksquare$ & $\phi$ & $\sim$ \\
    H2FA~\cite{xu2022h2fa} & 2022 & feat.+head & UDA(w) & $\leftrightarrow$ & $\triangle$ & $\circ$ & $\blacksquare$ & $\phi$ & $\sim$ \\
    Multi-task~\cite{zhang2022multiple} & 2022 & feat.+ regression & UDA & $\leftrightarrow$ & $\diamond$ & $\circ$ & $\blacksquare$ & $\phi$ & $\phi$ \\
    Task-align~\cite{zhao2022task} & 2022 & feat.+ regression & UDA & $\leftrightarrow$ & $\diamond$ & $\circ$ & $\blacksquare$ & $\phi$ & $\sim$ \\
    Li et al.~\cite{li2023distilling} & 2023 & feature & UDA & $\leftrightarrow$ & $\diamond$ & $\bullet$ & $\blacksquare$ & $\circ$ & $\sim$ \\
    Local-reg~\cite{piao2023unsupervised} & 2023 & regression & UDA & $\leftrightarrow$ & $\diamond$ & $\circ$ & $\blacksquare$ & $\phi$ & $\sim$ \\
    MARS~\cite{saoud2023mars} & 2023 & feature & DG & $\mathcal{R}$ & $\triangle$ & $\circ$ & $\blacksquare$ & $\phi$ & $\sim$ \\
    Biswas et al.~\cite{biswas2024domain} & 2024 & feature & UDA & $\leftrightarrow$ & $\diamond$ & $\circ$ & $\blacksquare$ & $\circ$ & $\sim$ \\
    Cai et al.~\cite{cai2024uncertainty} & 2024 & feature & UDA & $\leftrightarrow$ & $\triangle$ & $\bullet$ & $\blacksquare$ & $\Box$ & $\sim$ \\
    De-conf.~\cite{diamant2024confusing} & 2024 & head & s-free & $\leftrightarrow$ & $\triangle$ & $\circ$ & $\blacksquare$ & $\circ$ & $\sim$ \\
    He et al.~\cite{he2024recalling} & 2024 & head & UDA & $\leftrightarrow$ & $\triangle$ & $\bullet$ & $\square$ & $\circ$ & $\sim$ \\
    Align-Distill~\cite{kay2024align} & 2024 & feat.+head & UDA & $\leftrightarrow$ & $\triangle$ & $\circ$ & $\blacksquare$ & $\phi$ & $\sim$ \\
    Unbiased DG~\cite{liu2024unbiased} & 2024 & feature & DG & $\mathcal{R}$ & $\triangle$ & $\circ$ & $\blacksquare$ & $\phi$ & $\sim$ \\
    Song et al.~\cite{song2024cross} & 2024 & feature & UDA & $\leftrightarrow$ & $\triangle$ & $\circ$ & $\blacksquare$ & $\phi$ & $\sim$ \\
    Xu et al.~\cite{xu2024dst} & 2024 & feature & UDA & $\phi$ & $\triangle$ & $\circ$ & $\blacksquare$ & $\circ$ & $\rightarrow$ \\
    Pseudo ref.~\cite{zhang2024pseudo} & 2024 & head & UDA & $\leftrightarrow$ & $\triangle$ & $\circ$ & $\blacksquare$ & $\circ$ & $\sim$ \\
    Taming~\cite{zhao2024taming} & 2024 & feat.+head & UDA & $\leftrightarrow$ & $\triangle$ & $\bullet$ & $\phi$ & $\phi$ & $\sim$ \\
    DATR~\cite{chen2025datr} & 2025 & feat.+head & UDA & $\leftrightarrow$ & $\triangle$ & $\circ$ & $\blacksquare$ & $\circ$ & $\sim$ \\
    Gaussian~\cite{chen2025gaussian} & 2025 & feat.+head & UDA & $\leftrightarrow$ & $\triangle$ & $\bullet$ & $\blacksquare$ & $\circ$ & $\sim$ \\
    Refining~\cite{chen2025refining} & 2025 & feat.+head & UDA & $\leftrightarrow$ & $\triangle$ & $\circ$ & $\blacksquare$ & $\phi$ & $\sim$ \\
    WMFA~\cite{cheng2025wmfa} & 2025 & feature & UDA & $\leftrightarrow$ & $\triangle$ & $\circ$ & $\blacksquare$ & $\phi$ & $\sim$ \\
    Ge et al.~\cite{ge2025exploring} & 2025 & feat.+head & UDA & $\leftrightarrow$ & $\triangle$ & $\circ$ & $\blacksquare$ & $\phi$ & $\sim$ \\
    Differential~\cite{he2025differential} & 2025 & feature & UDA & $\leftrightarrow$ & $\triangle$ & $\circ$ & $\blacksquare$ & $\phi$ & $\sim$ \\
    Bridging labels~\cite{kennerley2025bridging} & 2025 & feature & UDA & $\leftrightarrow$ & $\triangle$ & $\bullet$ & $\phi$ & $\phi$ & $\rightarrow$ \\
    Large SSL~\cite{lavoie2025large} & 2025 & feature & UDA & $\leftrightarrow$ & $\triangle$ & $\circ$ & $\blacksquare$ & $\phi$ & $\sim$ \\
    A2MADA~\cite{li2025a2mada} & 2025 & feature & UDA & $\leftrightarrow$ & $\diamond$ & $\circ$ & $\blacksquare$ & $\Box$ & $\sim$ \\
    Dual-perspective~\cite{liu2025dual} & 2025 & feature & UDA & $\leftrightarrow$ & $\triangle$ & $\circ$ & $\blacksquare$ & $\phi$ & $\sim$ \\
    Semantic CLIP~\cite{liu2025semantic} & 2025 & feature & few-shot & $\phi$ & $\triangle$ & $\bullet$ & $\phi$ & $\circ$ & $\sim$ \\
    Mahayuddin et al.~\cite{mahayuddin2025lightweight} & 2025 & regression & UDA & $\leftrightarrow$ & $\diamond$ & $\circ$ & $\square$ & $\circ$ & $\sim$ \\
    CCLDet~\cite{shang2025ccldet} & 2025 & feature & UDA & $\leftrightarrow$ & $\triangle$ & $\circ$ & $\blacksquare$ & $\Box$ & $\sim$ \\
    VCR~\cite{vcr2025foundation} & 2025 & feat.+head & s-free & $\leftrightarrow$ & $\triangle$ & $\bullet$ & $\phi$ & $\circ$ & $\sim$ \\
    CMFAA-R-CNN~\cite{wang2025cross} & 2025 & feat.+head & UDA & $\leftrightarrow$ & $\triangle$ & $\circ$ & $\blacksquare$ & $\phi$ & $\sim$ \\
    M4-SAR~\cite{wang2025m4} & 2025 & feature & UDA & $\leftrightarrow$ & $\triangle$ & $\circ$ & $\blacksquare$ & $\phi$ & $\sim$ \\
    Unsupervised~\cite{wang2025unsupervised} & 2025 & feat.+head & UDA & $\leftrightarrow$ & $\triangle$ & $\circ$ & $\blacksquare$ & $\phi$ & $\sim$ \\
    Multi-scale~\cite{wei2025multi} & 2025 & feat.+head & UDA & $\leftrightarrow$ & $\diamond$ & $\circ$ & $\blacksquare$ & $\phi$ & $\sim$ \\
    FSDA-DETR~\cite{yang2025fsda} & 2025 & feature & few-shot & $\phi$ & $\triangle$ & $\circ$ & $\blacksquare$ & $\phi$ & $\sim$ \\
    Versatile~\cite{yang2025versatile} & 2025 & feat.+head & UDA & $\leftrightarrow$ & $\triangle$ & $\circ$ & $\blacksquare$ & $\phi$ & $\sim$ \\
    Source-free~\cite{yao2025source} & 2025 & feat.+head & s-free & $\leftrightarrow$ & $\triangle$ & $\circ$ & $\blacksquare$ & $\phi$ & $\sim$ \\
    FSDAOD~\cite{zhao2025fsdaod} & 2025 & feature & few-shot & $\phi$ & $\triangle$ & $\circ$ & $\blacksquare$ & $\phi$ & $\sim$ \\
    GAANet~\cite{zheng2025gaanet} & 2025 & feature & UDA & $\leftrightarrow$ & $\triangle$ & $\circ$ & $\blacksquare$ & $\phi$ & $\sim$ \\
    Universal~\cite{zheng2025universal} & 2025 & feat.+head & univ. & $\leftrightarrow$ & $\triangle$ & $\circ$ & universal & $\phi$ & $\sim$ \\
    HMDA-YOLO~\cite{zhu2025cross} & 2025 & feature & UDA & $\leftrightarrow$ & $\triangle$ & $\circ$ & $\blacksquare$ & $\phi$ & $\sim$ \\
    Geng et al.~\cite{geng2026cen} & 2026 & feature & DG & $\mathcal{R}$ & $\triangle$ & $\circ$ & $\blacksquare$ & $\phi$ & $\sim$ \\
   
\bottomrule
    \end{tabularx}
    \end{table*}
\FloatBarrier

\FloatBarrier
\begin{table}[t]
    \caption{Method-level diagnostic matrix showing whether each method explicitly addresses separation, recall, and calibration (ECE), with the observed dominant failure mode. Symbols: $\checkmark$ = addressed, $\times$ = not addressed, $\triangle$ = partial tendency, $\uparrow\!\mathcal{B}$ = bias amplification, $\mathcal{F}_{\text{reg}}\triangle$ = regression-focused partial effect, and $\tau$-sensitive = threshold-sensitive behavior.}
        \label{tab:diagnostic}
        \centering
        \scriptsize
        \setlength{\tabcolsep}{1.4pt}
    \renewcommand{\arraystretch}{0.95}
        % Vertical lines removed to match the table style used with booktabs.
        \begin{tabularx}{\columnwidth}{@{}>{\raggedright\arraybackslash}p{4cm}>{\centering\arraybackslash}p{2.3cm}>{\centering\arraybackslash}p{1.7 cm}>{\centering\arraybackslash}p{1.5 cm}>{\centering\arraybackslash}X@{}}
         \toprule
        \textbf{Method} & \textbf{Separation} & \textbf{Recall} & \textbf{ECE} & \textbf{Failure} \\
        \midrule
        DA-Faster \cite{chen2018domain} & $\checkmark$ & $\times$ & $\times$ & $\uparrow\!\mathcal{B}$ \\
        Strong-Weak \cite{saito2019strong} & $\checkmark$ & $\times$ & $\times$ & $\uparrow\!\mathcal{B}$ \\
        Yu et al. \cite{yu2019unsupervised} & $\checkmark$ & $\times$ & $\times$ & $\uparrow\!\mathcal{B}$ \\
        Category anchor \cite{zhang2019category} & $\checkmark$ & $\times$ & $\times$ & $\uparrow\!\mathcal{B}$ \\
        Selective \cite{zhu2019adapting} & $\checkmark$ & $\times$ & $\times$ & $\uparrow\!\mathcal{B}$ \\
        Nguyen et al. \cite{nguyen2020domain} & $\checkmark$ & $\times$ & $\times$ & $\uparrow\!\mathcal{B}$ \\
        Xu et al. \cite{xu2020cross} & $\checkmark$ & $\times$ & $\times$ & calibration focus $\triangle$ \\
        Zhang et al. \cite{zhang2020multi} & $\checkmark$ & $\times$ & $\times$ & $\uparrow\!\mathcal{B}$ \\
        CR-DA \cite{zheng2020cross} & $\checkmark$ & $\times$ & $\times$ & $\uparrow\!\mathcal{B}$ \\
        Scale DA \cite{chen2021scale} & $\checkmark$ & $\times$ & $\times$ & $\uparrow\!\mathcal{B}$ \\
        Unbiased MT \cite{deng2021unbiased} & $\checkmark$ & $\times$ & $\times$ & $\uparrow\!\mathcal{B}$ \\
        Mega-CDA \cite{vs2021mega} & $\checkmark$ & $\times$ & $\times$ & $\uparrow\!\mathcal{B}$ \\
        AFAN \cite{wang2021afan} & $\checkmark$ & $\times$ & $\times$ & $\uparrow\!\mathcal{B}$ \\
        C2FDA \cite{zhang2021c2fda} & $\checkmark$ & $\times$ & $\times$ & $\uparrow\!\mathcal{B}$ \\
        Cheng et al. \cite{cheng2022anchor} & $\checkmark$ & $\times$ & $\times$ & $\uparrow\!\mathcal{B}$ \\
        Do et al. \cite{do2022exploiting} & $\checkmark$ & $\times$ & $\times$ & $\uparrow\!\mathcal{B}$ \\
        Progressive \cite{gao2022progressive} & $\checkmark$ & $\times$ & $\times$ & $\uparrow\!\mathcal{B}$ \\
        Category contrast \cite{huang2022category} & $\checkmark$ & $\times$ & $\times$ & $\uparrow\!\mathcal{B}$ \\
        Dual instance \cite{jiao2022dual} & $\checkmark$ & $\times$ & $\times$ & $\uparrow\!\mathcal{B}$ \\
        S-DAYOLO \cite{li2022cross} & $\checkmark$ & $\times$ & $\times$ & $\uparrow\!\mathcal{B}$ \\
        SCAN \cite{li2022scan} & $\checkmark$ & $\times$ & $\times$ & $\uparrow\!\mathcal{B}$ \\
        H2FA \cite{xu2022h2fa} & $\checkmark$ & $\times$ & $\times$ & $\uparrow\!\mathcal{B}$ \\
        Multi-task \cite{zhang2022multiple} & $\checkmark$ & $\times$ & $\times$ & $\mathcal{F}_{\text{reg}}\triangle$ \\
        Task-align \cite{zhao2022task} & $\checkmark$ & $\times$ & $\times$ & $\mathcal{F}_{\text{reg}}\triangle$ \\
        Li et al. \cite{li2023distilling} & $\checkmark$ & $\times$ & $\times$ & calibration focus $\triangle$ \\
        Local regression \cite{piao2023unsupervised} & $\times$ & $\times$ & $\times$ & $\mathcal{F}_{\text{reg}}\triangle$ \\
        MARS \cite{saoud2023mars} & $\times$ & $\times$ & $\times$ & low mAP \\
        Biswas et al. \cite{biswas2024domain} & $\checkmark$ & $\times$ & $\times$ & $\uparrow\!\mathcal{B}$ \\
        Cai et al. \cite{cai2024uncertainty} & $\checkmark$ & $\times$ & $\times$ & calibration focus $\triangle$ \\
        De-confusing \cite{diamant2024confusing} & $\checkmark$ & $\times$ & $\times$ & $\uparrow\!\mathcal{B}$ \\
        He et al. \cite{he2024recalling} & $\times$ & $\times$ & $\times$ & $\tau$-sensitive \\
        Align-Distill \cite{kay2024align} & $\checkmark$ & $\times$ & $\times$ & $\uparrow\!\mathcal{B}$ \\
        Unbiased DG \cite{liu2024unbiased} & $\times$ & $\times$ & $\times$ & low mAP \\
        Song et al. \cite{song2024cross} & $\checkmark$ & $\times$ & $\times$ & $\uparrow\!\mathcal{B}$ \\
        Xu et al. \cite{xu2024dst} & $\checkmark$ & $\times$ & $\times$ & $\uparrow\!\mathcal{B}$ \\
        Pseudo-label refinement \cite{zhang2024pseudo} & $\checkmark$ & $\times$ & $\times$ & $\uparrow\!\mathcal{B}$ \\
        Taming \cite{zhao2024taming} & $\checkmark$ & $\times$ & $\times$ & calibration focus $\triangle$ \\
        DATR \cite{chen2025datr} & $\checkmark$ & $\times$ & $\times$ & $\uparrow\!\mathcal{B}$ \\
        Gaussian \cite{chen2025gaussian} & $\checkmark$ & $\times$ & $\checkmark$ & calibration focus $\triangle$ \\
        Refining \cite{chen2025refining} & $\checkmark$ & $\times$ & $\times$ & $\uparrow\!\mathcal{B}$ \\
        WMFA \cite{cheng2025wmfa} & $\checkmark$ & $\times$ & $\times$ & $\uparrow\!\mathcal{B}$ \\
        Ge et al. \cite{ge2025exploring} & $\checkmark$ & $\times$ & $\times$ & $\uparrow\!\mathcal{B}$ \\
        Differential \cite{he2025differential} & $\checkmark$ & $\times$ & $\times$ & $\uparrow\!\mathcal{B}$ \\
        Bridging labels \cite{kennerley2025bridging} & $\checkmark$ & $\times$ & $\times$ & $\uparrow\!\mathcal{B}$ \\
        Large SSL \cite{lavoie2025large} & $\checkmark$ & $\times$ & $\times$ & $\uparrow\!\mathcal{B}$ \\
        A2MADA \cite{li2025a2mada} & $\checkmark$ & $\times$ & $\times$ & $\uparrow\!\mathcal{B}$ \\
        Dual-perspective \cite{liu2025dual} & $\checkmark$ & $\times$ & $\times$ & $\uparrow\!\mathcal{B}$ \\
        Semantic CLIP \cite{liu2025semantic} & $\checkmark$ & $\times$ & $\times$ & few-shot $\triangle$ \\
        Mahayuddin et al. \cite{mahayuddin2025lightweight} & $\checkmark$ & $\times$ & $\times$ & $\tau$-sensitive \\
        CCLDet \cite{shang2025ccldet} & $\checkmark$ & $\times$ & $\times$ & $\uparrow\!\mathcal{B}$ \\
        VCR \cite{vcr2025foundation} & $\checkmark$ & $\times$ & $\times$ & $\uparrow\!\mathcal{B}$ \\
        CMFAA-R-CNN \cite{wang2025cross} & $\checkmark$ & $\times$ & $\times$ & $\uparrow\!\mathcal{B}$ \\
        M4-SAR \cite{wang2025m4} & $\checkmark$ & $\times$ & $\times$ & $\uparrow\!\mathcal{B}$ \\
        Unsupervised \cite{wang2025unsupervised} & $\checkmark$ & $\times$ & $\times$ & $\uparrow\!\mathcal{B}$ \\
        Multi-scale \cite{wei2025multi} & $\checkmark$ & $\times$ & $\times$ & $\uparrow\!\mathcal{B}$ \\
        FSDA-DETR \cite{yang2025fsda} & $\checkmark$ & $\times$ & $\times$ & $\uparrow\!\mathcal{B}$ \\
        Versatile \cite{yang2025versatile} & $\checkmark$ & $\times$ & $\times$ & $\uparrow\!\mathcal{B}$ \\
        Source-free \cite{yao2025source} & $\checkmark$ & $\times$ & $\times$ & $\uparrow\!\mathcal{B}$ \\
        FSDAOD \cite{zhao2025fsdaod} & $\checkmark$ & $\times$ & $\times$ & $\uparrow\!\mathcal{B}$ \\
        GAANet \cite{zheng2025gaanet} & $\checkmark$ & $\times$ & $\times$ & $\uparrow\!\mathcal{B}$ \\

        \bottomrule
        \end{tabularx}
    \end{table}
\FloatBarrier

% =============================================================================
\section{Discussion}
\label{sec:discussion}
The synthesis above identifies recurring patterns; this section clarifies why
those patterns matter for interpreting reported gains and for judging whether
they transfer to real deployments.

Table~\ref{tab:taxonomy-comparison} and Table~\ref{tab:diagnostic} indicate a
strong concentration of method choices: alignment-centric objectives, implicit
treatment of pipeline stages, and mostly closed-set assumptions. This
concentration is not only descriptive. It shapes what kinds of improvements are
easy to find, what failure modes remain hidden, and how confidently the field
can claim robust cross-domain generalization.

The first implication is \emph{benchmark-conditioned validity}. Many results
are obtained on controlled source-target pairs with stable label spaces and
predictable appearance shifts. In these settings, feature discrepancy is often
the dominant challenge, so alignment methods can show clear mAP gains. However,
deployment usually includes stronger shifts: semantic drift, long-tail classes,
scale and layout variation, unknown categories, and background reconfiguration.
As a result, gains should be interpreted as conditional rather than universal.
Performance on benchmark-favored settings does not, by itself, establish robust
transfer under broader shift families.

The second implication is \emph{evaluation ambiguity}. mAP remains the dominant
summary metric, but it mixes proposal quality, classification, localization, and
confidence behavior into one number. Similar mAP improvements can arise from
different mechanisms, and those mechanisms may have very different deployment
risk profiles \cite{zhao2022task,zhang2022multiple,he2025differential}. Without
stage-wise diagnostics, it is difficult to attribute where a method helps, why
it helps, and whether it can be reliably composed with other methods.

The third implication is \emph{objective mismatch between research and
deployment}. Real systems operate under asymmetric costs: missing critical
objects, producing unstable confidence, or degrading localization under shift
can be far more costly than small average changes in aggregate accuracy. A
single mAP value cannot capture these asymmetries. This is one reason why
progress measured on benchmark leaderboards can overstate real robustness.

A related issue is \emph{alignment dominance}. Alignment is attractive because
it is general, implementation-friendly, and often effective on standard
benchmarks. But when one paradigm dominates both method design and evaluation,
the evidence base can become self-reinforcing. Alternatives such as explicit
proposal adaptation, calibration-aware objectives, causal invariance, and
uncertainty-constrained learning remain comparatively under-validated
\cite{liu2024unbiased,geng2026cen}. The point is not that alignment fails, but
that methodological diversity is still insufficient for strong falsification.

Overall, a more reliable notion of progress should combine aggregate accuracy
with mechanism-level evidence: stage-wise diagnostics, calibration reporting,
and evaluation across heterogeneous shift types. Under this lens, CDOD research
can move from benchmark-specific gains toward findings that are transportable
across domains and dependable in practice.

% =============================================================================
% =============================================================================

\section{Datasets, Benchmarks, and Evaluation Protocols}
\label{sec:datasets}

Datasets and evaluation protocols strongly shape what we learn about CDOD
methods. This section reviews the most common benchmark datasets and discusses
what kinds of domain shift they represent, what they test effectively, and
where their limitations lie.

Most CDOD benchmarks are built by pairing datasets with different data
distributions. The gap may come from environment, capture conditions, sensor
configuration, or annotation policy. Table~\ref{tab:cdod_datasets} summarizes
the datasets most often used in practice, their typical source/target roles,
and the dominant shift each one introduces.

\begin{table*}[t]
    \centering
    \footnotesize
    \caption{Common datasets used in CDOD benchmarks, summarizing modality, scale, annotation volume, typical role, and dominant shift type. Acronyms: S = Source, T = Target. Symbol: $\sim$ indicates approximate counts.}
    \label{tab:cdod_datasets}
    \setlength{\tabcolsep}{1pt}
    \begin{tabularx}{\textwidth}{@{}l c c c c c c X@{}}
    \toprule
    \textbf{Dataset} & \textbf{Year} & \textbf{Modality} & \textbf{\#Images} & \textbf{\#Cls} & \textbf{\#Anno} & \textbf{Role} & \textbf{Domain Shift} \\
    \midrule
    PASCAL VOC \cite{everingham2010pascal} & 2007--2012 & RGB & $\sim$16.5K & $\sim$20 & $\sim$40K & S/T & mild scene shift \\
    MS COCO \cite{lin2014coco} & 2014 & RGB & $\sim$330K & $\sim$80 & $\sim$2.5M & S & scene diversity \\
    ImageNet DET \cite{ILSVRC15} & 2013 & RGB & $\sim$450K & $\sim$200 & $\sim$500K & S & fine-grained category \\
    \midrule
    Cityscapes \cite{cordts2016cityscapes} & 2016 & RGB & $\sim$3.0K & $\sim$8 & $\sim$65K & T & urban scene shift \\
    Foggy Cityscapes \cite{sakaridis2018semantic} & 2018 & RGB & $\sim$3.0K & $\sim$8 & $\sim$65K & T & weather (clear$\rightarrow$fog) \\
    SIM10K \cite{richter2016playing} & 2018 & RGB (Synthetic) & $\sim$10K & $\sim$1 & $\sim$58K & S & synth$\rightarrow$real \\
    GTA5 \cite{johnson2016driving} & 2016 & RGB (Synthetic) & $\sim$25K & $\sim$9 & $\sim$300K & S & synth$\rightarrow$real \\
    SYNTHIA \cite{ros2016synthia} & 2016 & RGB (Synthetic) & $\sim$9.4K & $\sim$9 & $\sim$200K & S & synth$\rightarrow$real \\
    \midrule
    BDD100K \cite{yu2020bdd100k} & 2020 & RGB / Video & $\sim$100K & $\sim$10 & $\sim$1.8M & S/T & scene/weather/light \\
    Dark Zurich \cite{sakaridis2019guided} & 2020 & RGB & $\sim$3K & $\sim$8 & $\sim$40K & T & day$\rightarrow$night \\
    \midrule
    KITTI \cite{Geiger2013IJRR} & 2012 & RGB + LiDAR & $\sim$15K & $\sim$3 & $\sim$80K & S & sensor shift \\
    nuScenes \cite{caesar2020nuscenes} & 2019 & RGB + LiDAR & $\sim$1.4M frames & $\sim$10 & $\sim$1.4M & T & sensor/scene shift \\
    Waymo Open \cite{sun2020scalability} & 2020 & RGB + LiDAR & $\sim$12M & $\sim$4 & $\sim$10M & T & sensor-scale shift \\
    \bottomrule
    \end{tabularx}
    \end{table*}

The scientific value of these benchmarks depends not only on dataset size, but
on the \emph{kind} of shift they induce. PASCAL VOC
\cite{everingham2010pascal} remains useful as a relatively clean, smaller-scale
benchmark with stable classes and moderate scene complexity, which helps with
controlled diagnostics and overfitting checks. MS COCO \cite{lin2014coco}
contributes strong intra-class variation (pose, context, scale, clutter, and
crowding), making it a useful stress test for representation robustness and
long-tail behavior. ImageNet DET \cite{ILSVRC15} adds fine-grained semantic
diversity, where transfer often exposes boundary ambiguity and
class-conditional mismatch.

Cityscapes \cite{cordts2016cityscapes} is valuable because it is structured yet
distributionally narrow: viewpoint, geometry, and object layout are consistent
enough that even small contextual changes are visible under transfer. Foggy
Cityscapes \cite{sakaridis2018semantic} introduces realistic visibility
degradation (contrast attenuation, blur, and loss of distant detail), making it
effective for probing weather-driven localization and confidence drift. Dark
Zurich \cite{sakaridis2019guided} is similarly important for illumination shift:
low light, sensor noise, and high dynamic range effects challenge both
foreground separation and confidence calibration in day-to-night transfer.

SIM10K \cite{richter2016playing}, GTA5 \cite{johnson2016driving}, and SYNTHIA
\cite{ros2016synthia} remain central for synthetic-to-real studies with
different realism levels and label scopes. Their strength is controlled
annotation quality combined with realistic rendering gaps, including texture
bias, lighting mismatch, material differences, and simulator-specific priors.
SIM10K is especially informative in class-limited transfer (often car-centric),
where improvements can be strong but narrow. GTA5 and SYNTHIA provide broader
urban diversity, making them stronger tests of whether adaptation learns
transferable structure rather than simulator artifacts.

BDD100K \cite{yu2020bdd100k} is one of the most informative real-world CDOD
datasets because it combines scene, weather, and illumination variability at
scale (day/night, clear/rain/fog, highway/city/residential), with strong
temporal diversity from video. This creates interacting shifts that are closer
to deployment than single-factor perturbations. KITTI \cite{Geiger2013IJRR},
nuScenes \cite{caesar2020nuscenes}, and Waymo Open \cite{sun2020scalability}
extend the challenge to multimodal sensing, where resolution, field-of-view,
camera--LiDAR calibration, motion blur, range sparsity, and annotation protocol
differences can dominate over appearance shift. These settings test whether a
method adapts geometric behavior and proposal quality, not only feature space.

Taken together, the datasets are complementary. Smaller clean sets (for example,
PASCAL VOC) support mechanism-level analysis; synthetic sources (SIM10K, GTA5,
SYNTHIA) provide controlled but biased supervision; adverse-condition targets
(Foggy Cityscapes, Dark Zurich) stress weather and illumination robustness; and
large heterogeneous corpora (MS COCO, BDD100K, nuScenes, Waymo Open) better
reflect deployment diversity. Robust CDOD claims should therefore be supported
by evaluation across multiple shift types, not a single benchmark pair.

% =============================================================================
% =============================================================================

\section{Deep Challenges in Cross-Domain Object Detection}
\label{sec:deep-challenges}

Standard narratives attribute CDOD difficulty to a ``large domain gap'' or ``lack of labeled target data'' \cite{chen2018domain,zhu2019adapting,saito2019strong}. Those factors matter, but they obscure subtler challenges that arise from the structure of the detection task and the behavior of adaptation mechanisms \cite{zhao2022task,zhang2022multiple}. This section examines \textbf{seven} core challenges that determine when and why adaptation fails; the rest (scale, NMS, label noise) are noted briefly where relevant.

% ---------------------------------------
\subsection{Entanglement Between Localization and Classification Under Shift}
% ---------------------------------------

Detection requires both a class label and a bounding box \cite{chen2018domain,zhao2022task}. The two outputs share the same backbone and often the same neck; they are optimized with a single loss that sums classification and regression terms. Under domain shift, however, the \emph{causes} of degradation differ: classification may fail due to semantic or style confusion, while localization may fail due to shifts in feature scale, object scale, or the distribution of proposal locations \cite{zhao2022task,zhang2022multiple}. The shared representation creates \emph{entanglement}: gradients from the classification loss and the regression loss flow back through the same layers. Aligning features to improve classification for example, via domain confusion can alter the feature scale or the spatial structure that the regression head relies on, and vice versa \cite{zhu2019adapting,he2025differential}. Disentangling the two under shift is difficult because there is no clean separation in the architecture--no guarantee that a change that helps one head does not hurt the other. Task-specific alignment for example, separate discriminators or losses for classification and regression \cite{zhao2022task,zhang2022multiple,he2025differential} is a partial remedy but is not yet standard; moreover, it does not remove the shared representation, so entanglement remains at the feature level. The challenge is thus not only to adapt both heads but to do so in a way that does not set them in opposition \cite{zhao2022task,he2025differential,zhou2025ccanet}. This is a structural challenge that does not arise in classification-only domain adaptation \cite{chen2018domain}.

% ---------------------------------------
\subsection{Proposal Instability Across Domains}
% ---------------------------------------

In two-stage detectors \cite{chen2018domain,saito2019strong}, the RPN or proposal module is trained on source data and must generalize to the target. Proposals carry objectness and their distribution (number, scale, aspect ratio, overlap with GT) is domain-dependent. Under shift, the RPN can produce too few proposals (missing detections), too many low-quality ones, or proposals biased in location or \emph{scale}--object scale varies across domains (surveillance vs.\ drone, resolution, FOV), and scale shift directly affects proposal coverage \cite{chen2021scale,li2025seen}. This \emph{proposal instability} is rarely a first-class object of adaptation. Most alignment methods operate \emph{after} proposals; they assume proposal quality transfers. When $P_T(b|x)$ is very different for example, different object sizes or clutter, downstream alignment cannot recover missed objects (Observation 1). One-stage and query-based detectors face the same issue: anchors or queries are source-tuned. The challenge is to stabilize or adapt the proposal mechanism without target boxes \cite{saito2019strong,li2025seen,li2023learning}.

% ---------------------------------------
\subsection{Calibration Under Domain Mismatch}
% ---------------------------------------

A detector is \emph{calibrated} when its confidence scores reflect the actual probability of correctness for example, when a prediction with score 0.8 is correct about 80\% of the time. Calibration is typically studied in-domain \cite{cai2024uncertainty}; under domain shift, it often breaks. The model may be overconfident on target data (high scores for wrong or sloppy detections) or underconfident (low scores for correct detections), and the miscalibration can vary by class, scale, or region. This matters for CDOD because many adaptation strategies \emph{rely on confidence}: pseudo-label selection by threshold, uncertainty-weighted alignment, and curriculum learning \cite{saito2019strong,yang2025versatile,wei2025multi,chen2025gaussian} all assume that confidence is a usable proxy for correctness. When calibration degrades in the target domain, high-confidence pseudo-labels can be wrong (amplifying false positives) and low-confidence correct detections can be discarded (reducing recall). Calibration under domain mismatch is not the same as ``domain gap''--it is a property of the output distribution of the model, and it can degrade even when feature alignment is successful. Recalibrating without target labels is difficult; temperature scaling and related techniques assume a validation set from the same distribution \cite{cai2024uncertainty}. The challenge is to maintain or restore calibration during adaptation \cite{chen2025gaussian}, or to design adaptation mechanisms that do not depend critically on well-calibrated confidence.

% ---------------------------------------
\subsection{Background Shift vs.\ Foreground Shift Asymmetry}
% ---------------------------------------

Domain shift affects both foreground (objects of interest) and background (everything else). The two need not shift in the same way: the target may have similar objects but very different backgrounds for example, same classes in a new city or new sensor, or similar backgrounds but different object appearance. Most alignment methods do not distinguish the two. Image-level or global feature alignment treats the whole image as one unit and can \emph{over-align background} while under-aligning foreground, or vice versa \cite{zhu2019adapting,do2022exploiting}. When background dominates the image (as it often does), alignment can be driven mainly by background statistics, so that foreground features are pulled toward a background-influenced mean and discriminativity drops. Foreground-focused or instance-level alignment is a partial remedy \cite{jiao2022dual,zhu2019adapting} but requires a notion of ``foreground'' in the target domain without labels--e.g., via objectness, attention, or propagation from source. The asymmetry is that background is abundant and easy to align (large, consistent regions), while foreground is sparse and heterogeneous; yet detection performance depends on foreground \cite{zhu2019adapting,jiao2022dual,do2022exploiting}. The challenge is to design alignment that is \emph{foreground-aware} or that down-weights background so that foreground structure is preserved or explicitly aligned \cite{vs2021mega,he2025differential}. This goes beyond ``closing the gap''--it requires a structural choice about what to align and what to protect.

% ---------------------------------------
\subsection{Open-Set and Category-Conditional Misalignment}
% ---------------------------------------
Even when marginal feature distributions are aligned, the \emph{conditional} per class can remain misaligned--some classes transfer well, others do not \cite{zhu2019adapting,vs2021mega}. Category-agnostic alignment can average over the discrepancy and leave the worst classes under-adapted. \emph{Open-set} \cite{zheng2025universal}: when the target has classes not in the source, closed-set methods force target-private instances into source clusters (negative transfer). Category-aware alignment \cite{vs2021mega,jiang2025adaptive} helps but requires target-class assignment without labels (pseudo-labels are noisy). The challenge is category-conditional alignment and robust separation of shared vs.\ private classes without labels \cite{zheng2025universal,pan2020exploring}; both are fragile in practice \cite{vs2021mega,jiang2025adaptive}.

% ---------------------------------------
\subsection{False Positive Amplification in Pseudo-Labeling}
% ---------------------------------------

Self-training and pseudo-labeling use the model's own predictions on the target domain as supervision. When the model makes systematic errors for example, confuses background with a frequent class, or produces duplicate boxes, those errors become pseudo-labels and are reinforced in the next round of training \cite{saito2019strong,wang2025unsupervised,chen2025refining,wei2025multi}. \emph{False positives} are particularly dangerous: a high-confidence wrong detection is likely to be selected as a pseudo-label and then learned as correct. Over iterations, the model can become more and more confident on a growing set of false positives, especially if the threshold for accepting pseudo-labels is not conservative enough or if there is no mechanism to correct mistakes. This is not simply ``noisy labels''--it is a feedback loop in which error begets error. Mitigations for example, confidence thresholds, teacher-student with EMA, filtering with VLMs \cite{kim2024vlm,chen2025refining} can reduce but not eliminate the risk; as long as the model is the sole source of target labels, some false positives will be accepted. The challenge is to break the amplification loop: to incorporate an external signal for example, foundation models \cite{vcr2025foundation,wu2023clipself}, contrastive objectives \cite{jia2025contrastive} or to design selection and weighting schemes that are robust to the model's own bias. This is a deep challenge because it is inherent to the self-training paradigm, not just to a particular method \cite{saito2019strong,yang2025versatile,wang2025unsupervised,chen2025refining}.

% ---------------------------------------
\subsection{Long-Tail Domain Shift}
% ---------------------------------------

In many applications, the \emph{distribution of domains} is long-tailed: a few ``head'' domains for example, common weather, common sensors have abundant data, while many ``tail'' domains (rare weather, rare locations, new sensors) have little or no data. Adaptation is typically studied in the setting of one source and one target \cite{chen2018domain,zheng2020cross,saito2019strong}; in practice, a single model may need to perform across many tail domains \cite{liu2024unbiased,zhao2025few,yang2025fsda}. Tail domains are difficult because there is little target data to align to, and methods that rely on target statistics for example, prototype updates, batch normalization statistics can be unstable \cite{jiang2025adaptive,zhao2025few}. Moreover, tail domains may be underrepresented in any pretraining or foundation model, so that external priors are less reliable \cite{wu2023clipself,vcr2025foundation}. Long-tail domain shift is thus not only ``few-shot per domain'' but a structural property of the deployment distribution: the model must generalize to domains that are rare and diverse. Current benchmarks rarely evaluate on many target domains or on a long-tail of domain types \cite{liu2024unbiased,geng2026cen}; the challenge is to design adaptation that is sample-efficient per domain \cite{zhao2025few,yang2025fsda} and that does not forget or degrade on head domains when adapting to tail domains.

\textit{Other challenges}: NMS and label noise propagate through the pipeline \cite{zhang2019category,inoue2018cross}. Fixed NMS tuned on source can over- or under-suppress on target; source label noise and annotation protocol shift can compound with pseudo-label bias. We do not expand these here; they reinforce that adaptation is pipeline-wide.

% ---------------------------------------
\subsection{Synthesis}
% ---------------------------------------

The seven core challenges are interrelated and structural--not reducible to ``large domain gap'' or ``no labels.'' The field has overfit to synthetic-to-real where many challenges are mild \cite{li2022cross,chen2018domain}; gains often do not generalize to real-to-real, open-set, or long-tail \cite{wang2025sr,liang2025perspective,wang2025unsupervised}. A mature methodology would address them explicitly (disentangle localization and classification, stabilize proposals, maintain calibration, align foreground and category-conditionally, control pseudo-label noise) and adopt stage-wise metrics (Sec.~\ref{sec:operationalizing}).

% =============================================================================
% =============================================================================

\section{Failure Modes in Current CDOD Methods}
\label{sec:failure-modes}

Five failure modes recur \cite{chen2018domain,zhu2019adapting,saito2019strong,zhao2022task}. They are not edge cases but consequences of how the field works: alignment-centric, closed-set, synthetic-to-real-heavy, mAP-only evaluation \cite{zheng2020cross,liu2024unbiased}. Our goal is to expose the underlying mechanisms behind these failures rather than merely cataloguing their symptoms.

% ---------------------------------------
\subsection{Why Adversarial Alignment Often Hurts Rare Classes}
% ---------------------------------------

Alignment minimizes discrepancy on the \emph{marginal} feature distribution \cite{zhu2019adapting,he2025differential}; the gradient is strongest where marginals differ most. Frequent classes and background dominate; rare classes contribute few samples, so their distribution is under-specified \cite{chen2018domain,zhu2019adapting}. The discriminator has little incentive to align them; pulling rare-class features toward the majority mean can \emph{reduce} discrepancy. Alignment compresses or collapses the rare-class subspace. Category-aware alignment \cite{vs2021mega} needs correct target-class assignment; with pseudo-labels, rare classes have the noisiest assignments--the remedy is fragile. Structural: marginal alignment optimizes for the majority at the tail's expense \cite{chen2018domain,zhu2019adapting,vs2021mega,he2025differential}.

% ---------------------------------------
\subsection{Why Self-Training Amplifies Bias}
% ---------------------------------------

Self-training \cite{saito2019strong,yang2025versatile,wang2025unsupervised} uses the model's target predictions as pseudo-labels. The model is source-biased; biased predictions are selected by confidence and used as supervision. The next iteration reproduces them; confidence on the biased set rises; more of the same is admitted. No corrective signal--only the model's output. Confirmation bias is built in. Teacher-student and EMA slow drift but do not remove the loop \cite{yang2025versatile,zhao2024taming}; they assume the teacher stabilizes near the true distribution, which need not hold when source and target conditionals differ \cite{saito2019strong}. Amplification: closed-loop learning with a single, biased labeler \cite{saito2019strong,chen2025refining,wei2025multi}.

% ---------------------------------------
\subsection{Why Synthetic-to-Real Gains Do Not Transfer to Real-to-Real}
% ---------------------------------------

Synthetic-to-real \cite{chen2018domain,zheng2020cross,li2022cross}: controlled covariate shift (appearance); alignment of statistics addresses it. Real-to-real mixes covariate, semantic (class mix, context), and contextual (scale, density, layout) change \cite{wang2025sr,wang2025unsupervised,liang2025perspective}. Methods tuned on synthetic-to-real assume closed set and similar layout; on real-to-real, the same alignment can pull apart features that should stay separate or leave important shifts unaddressed. Synthetic-to-real is also easier--the gap is obvious, alignment yields visible gains \cite{chen2018domain,zheng2020cross}; real-to-real gaps are subtler, alignment can over-smooth or under-fit \cite{wang2025sr,liang2025perspective,wang2025unsupervised}. Transfer fails: shift type and objective are mismatched \cite{li2022cross,wang2025sr}.

% ---------------------------------------
\subsection{Why Feature Alignment Ignores Detection Geometry}
% ---------------------------------------

Feature alignment is on vectors; it has no notion of box, scale, or layout \cite{chen2018domain,zhu2019adapting}. The loss matches distributions in value space, not \emph{where} features sit or the \emph{scale} of regression targets. When alignment distorts magnitude or correlation for example, for the discriminator, the regression head receives a different input distribution and produces biased or high-variance boxes \cite{zhao2022task,zhang2022multiple}. No alignment term penalizes this; geometry is a side effect, not a constraint. Task-specific or bin-wise localization alignment \cite{zhao2022task,he2025differential} exists but is the exception. Failure by design: the objective is distribution alignment, not geometric consistency \cite{chen2018domain,zhu2019adapting,zhao2022task,zhang2022multiple}.

% ---------------------------------------
\subsection{Applying the Framework to Representative Methods}
% ---------------------------------------
To illustrate how Eq.~\ref{eq:grand-unifying} and Eq.~\ref{eq:decomposition} expose method limitations, we analyze four representative approaches.

\begin{figure}[t]
\centering
\includegraphics[width=\columnwidth]{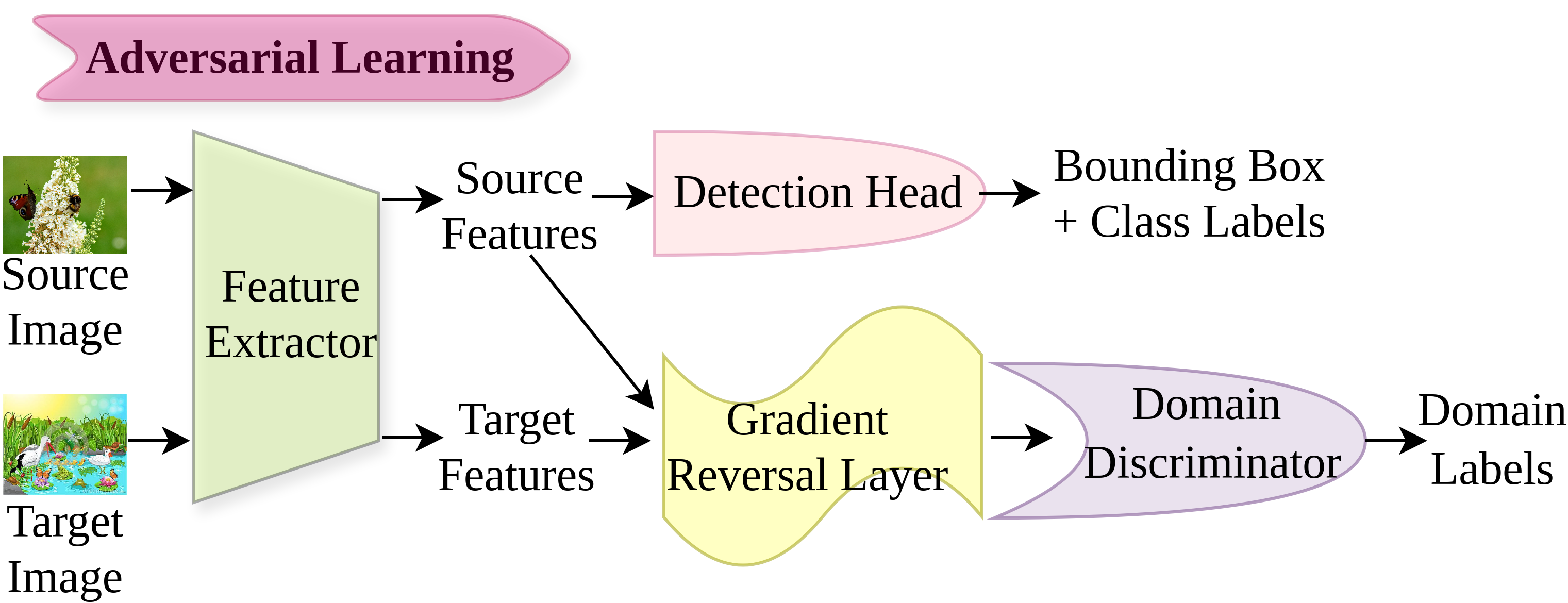}
\caption{Adversarial domain adaptation for CDOD: a domain discriminator drives feature alignment between source and target.}
\label{fig:adversarial}
\end{figure}

\emph{Adversarial domain adaptation} for example, DA-Faster \cite{chen2018domain}, selective cross-domain alignment \cite{zhu2019adapting}, differential alignment \cite{he2025differential}. \emph{Preserved:} Feature alignment via gradient reversal or domain discriminator enforces similar marginal distributions of backbone features $\phi(x)$ on source and target see fig. \ref{fig:adversarial}, partially addressing discriminativity $\mathrm{Sep}$ by reducing domain-specific structure, but only at the marginal level. \emph{Ignored:} (1) Proposal coverage: adversarial methods do not explicitly constrain proposal recall; they assume feature alignment indirectly preserves proposal quality. When $P_T(b|x)$ degrades for example, different object scales or clutter, the method has no direct mechanism. (2) Calibration: no loss term targets calibration; confidence can drift on target. \emph{Failure:} Rare classes suffer (Sec.~\ref{sec:failure-modes}) because marginal alignment favors majority classes and background; rare-class features are under-specified and can collapse. Geometry is ignored: the discriminator has no notion of box or scale, so regression receives mis-scaled features. Under Eq.~\ref{eq:grand-unifying}, adversarial methods preserve at most one invariant and leave proposal coverage and calibration unconstrained.

\begin{figure}[t]
\centering
\includegraphics[width=\columnwidth]{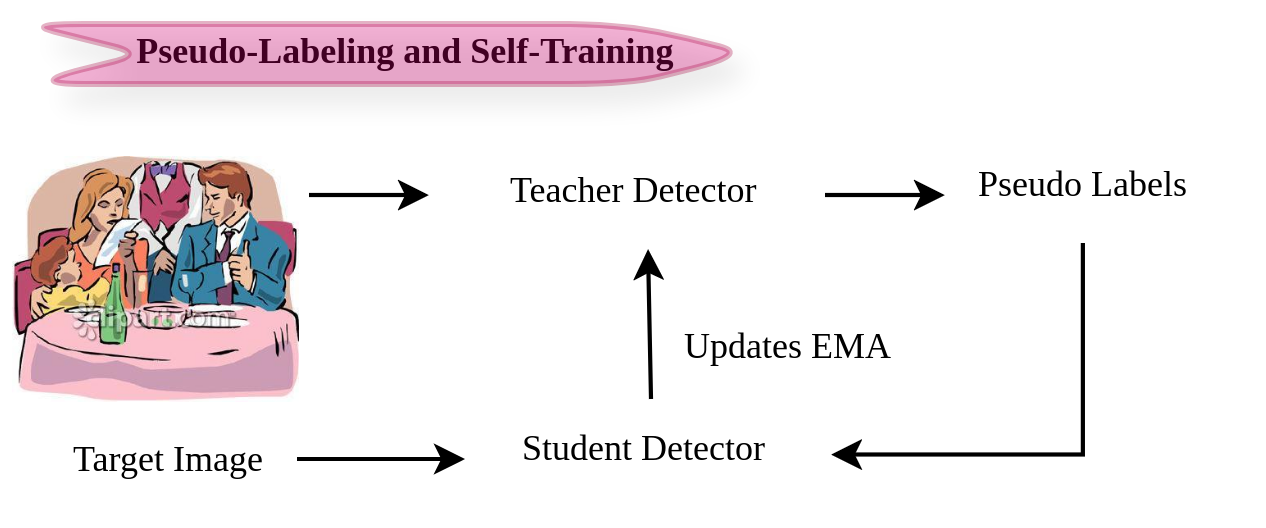}
\caption{Pseudo-label self-training in CDOD: a teacher generates target pseudo-labels to supervise a student, with risk of confirmation bias.}
\label{fig:pseudolabel}
\end{figure}

\emph{Self-training} for example, Strong-Weak \cite{saito2019strong}, Versatile Teacher \cite{yang2025versatile}. \emph{Preserved:} Target-side supervisory signal via pseudo-labels can improve discriminativity $\mathrm{Sep}$ and recall if the teacher model is well-calibrated see fig. \ref{fig:pseudolabel}. \emph{Ignored:} Calibration is used for pseudo-label selection but not explicitly constrained; proposal recall is not explicitly maintained. \emph{Failure:} False positive amplification and confirmation bias occur (Table~\ref{tab:diagnostic}) because the model's own biased predictions become supervision. When $P_T(b|x)$ is biased and pseudo-labels lock onto errors, the feedback loop reinforces mistakes. Teacher-student and EMA slow drift but do not remove the confirmation loop.

\emph{Domain generalization} for example, Unbiased DG \cite{liu2024unbiased}. \emph{Preserved:} Robustness over a family of domains is achieved through augmentation and style diversification; no target data is required at training, so the method generalizes to unseen domains. \emph{Ignored:} Target-specific recall and calibration cannot be optimized without target data at training time. \emph{Failure:} Lower mAP on specific target domains is structural, not a bug; domain generalization trades off in-domain performance for out-of-domain robustness. The method cannot adapt to target-specific characteristics.

\emph{Universal domain adaptation} for example, \cite{zheng2025universal}. \emph{Preserved:} Open-set and partial-set scenarios are handled by identifying shared versus private classes through clustering or thresholding mechanisms. \emph{Ignored:} Proposal recall and calibration are not explicitly maintained; the method relies on thresholds and clustering to identify private classes without target labels. \emph{Failure:} Threshold sensitivity and fragile identification of shared versus private classes occur because separation relies on heuristics without labeled target data. The method struggles when class boundaries are ambiguous or when target-private classes are similar to source classes (Table~\ref{tab:diagnostic}).

% ---------------------------------------
\subsection{Why Current Methods Fail Under Open-Set Conditions}
% ---------------------------------------

Closed-set methods push target features toward source-class structure; pseudo-labels assign target instances to source classes \cite{saito2019strong,yang2025versatile}. When the target has \emph{private} classes, their instances are forced into source clusters or background--negative transfer \cite{pan2020exploring}. Open-set and universal DAOD \cite{zheng2025universal} try to identify shared vs.\ private (thresholds, clustering, auxiliary models) but without target labels identification is unreliable. Current methods are closed-set by construction: no ``unknown'' in the label space or objective; the training signal pulls target data into source categories. Handling open-set would require explicit unknown class, robust separation of shared vs.\ private without labels, and an objective that does not align private-class features to source \cite{zheng2025universal,pan2020exploring}; few methods satisfy all three \cite{zheng2025universal}.

% =============================================================================
% =============================================================================

\section{Future Directions}
\label{sec:future-directions}

Problem: high-impact directions remain underdeveloped because current practice overfits to synthetic-to-real benchmarks \cite{chen2018domain,zheng2020cross}, relies heavily on alignment narratives \cite{zhu2019adapting,he2025differential}, and evaluates mostly with mAP-only reporting \cite{saito2019strong}. Direction: the nine subsections below frame the missing bottleneck and outline concrete research questions aimed at changing objectives, adaptation protocols, and evaluation so the required invariants are preserved under stronger shift.

% ---------------------------------------
\subsection{Causal Modeling of Domain Shift}
% ---------------------------------------

Problem: CDOD rarely operationalizes causal invariants, so ``style vs.\ content'' remains narrative rather than actionable. Causal modeling targets \emph{causal} invariants (style vs.\ content) across all stages, rather than correlational alignment. Correlational alignment can match statistics that are spuriously associated with domain for example, background color, leading to failure under new shifts \cite{zhu2019adapting,do2022exploiting}. Causal framing separates stable (causal) from unstable (spurious) structure and can in principle guarantee transfer when only the right variables for example, style change \cite{xu2024dst,tulu2025wct}. However, almost no CDOD work specifies a causal graph for example, style $\rightarrow$ image, content $\rightarrow$ label, performs intervention, or evaluates under intervention. Key research questions include: What causal graph for detection (images, boxes, labels, domain) is identifiable from observational source and target data \cite{xu2024dst}? Can intervention on style for example, via generation yield estimators that transfer under style shift \cite{tulu2025wct,feng2025vision}? How to evaluate causal CDOD (intervention-based protocols)?

% ---------------------------------------
\subsection{Foundation Models for CDOD}
% ---------------------------------------
\begin{figure}[t]
\centering
\includegraphics[width=0.5\columnwidth]{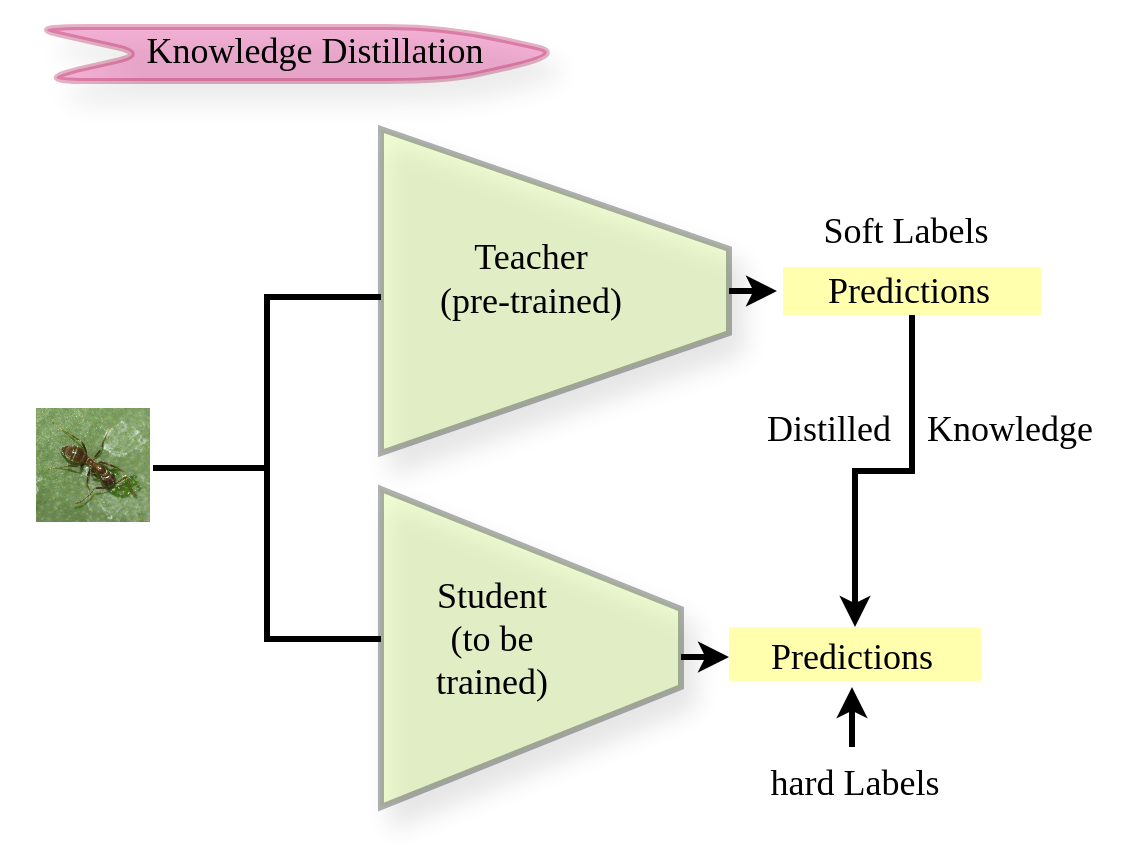}
\caption{Distillation for CDOD: transferring teacher knowledge for example, foundation-model or EMA teacher signals to improve target robustness.}
\label{fig:distillation}
\end{figure}
Distillation-style teacher signals (Fig.~\ref{fig:distillation}) can strengthen target-side guidance when pseudo-label and confidence mechanisms break under shift. Problem: reliable target-side signals are weak under shift, so pseudo-label and confidence mechanisms can break. Foundation models primarily affect feature and proposal stages (via pseudo-labels and objectness priors) and can support calibration, preserving discriminativity through better target-side signals. Large vision and vision-language models \cite{vcr2025foundation,wu2023clipself,li2023distilling} provide better pseudo-labels, objectness priors, and feature targets without target labels. They can break the self-training confirmation loop and supply calibration or open-vocabulary signals that detectors lack. However, integration is ad hoc: VLMs for filtering \cite{kim2024vlm}, DINO for feature alignment, SAM for foreground. There is no unified view of how to use foundation models as \emph{adaptation primitives} (labelers, regularizers, or teachers) across the pipeline \cite{wu2025dara}, nor when they fail for example, target far from pretraining distribution. Research questions include: How to combine foundation-model signals with detector adaptation at feature, proposal, and head stages \cite{vcr2025foundation,wu2023clipself,li2023distilling}? When do foundation-model priors hurt (negative transfer) \cite{wu2025dara}? Can small, efficient detectors be adapted using frozen foundation encoders without blowing compute \cite{lavoie2025large}?

% ---------------------------------------
\subsection{Test-Time Adaptation for Detection}
% ---------------------------------------

Problem: shift is often only observed at inference, but detection adaptation is usually designed offline. Test-time adaptation operates across all stages (normalization, adapters at test time), preserving or restoring calibration and proposal coverage when shift is observed at inference. Deployment often encounters shift only at test time (new camera, new location) \cite{wang2025v2x,luo2025mas}. Batch adaptation with target data is not always possible; per-image or per-batch adaptation at inference would allow continuous adaptation without retraining \cite{liu2025adaptive}. However, test-time adaptation (TTA) is established in classification but barely explored in detection \cite{liu2025adaptive}. Detection has two outputs, proposals, and NMS--all of which may need to adapt. Memory and latency constraints are tighter at test time. Key research questions include: What detector parameters can be adapted at test time for example, normalization, small adapters without catastrophic forgetting? How to obtain a test-time objective for detection without labels for example, consistency, entropy, foundation-model similarity? How to avoid collapse when adapting on a single or few test images?

% ---------------------------------------
\subsection{Continual Domain Adaptation}
% ---------------------------------------

Problem: real deployments encounter streams of new domains, but CDOD is usually treated as one-off transfer. Continual domain adaptation operates across all stages, preserving performance on prior domains (stability) while adapting to new ones (plasticity). In practice, a model may see a stream of domains for example, new cities, new seasons, while most CDOD studies still assume a single source-target transfer episode \cite{chen2018domain,saito2019strong,zheng2020cross}. Adapting to each new domain from scratch is costly; adapting sequentially risks forgetting previous domains. Continual domain adaptation (CDA) aims to accumulate and retain knowledge across domains. However, CDA for detection is still weakly benchmarked in current CDOD protocols \cite{liu2024unbiased,geng2026cen}. Most work is single source $\rightarrow$ single target. Replay, regularization, and parameter isolation from continual learning are not yet standard in CDOD; the interaction between domain shift and catastrophic forgetting is underexplored. Research questions include: How to adapt a detector to domain $t+1$ while retaining performance on domains $1,\ldots,t$ without storing past data? Can replay or distillation from a small buffer of past-domain statistics suffice? How to define and evaluate ``stability'' vs.\ ``plasticity'' across domains for detection?

% ---------------------------------------
\subsection{Domain Generalization Without Explicit Adaptation}
% ---------------------------------------

Problem: when no target data is available, many adaptation approaches cannot be applied, yet DG for detection remains relatively underdeveloped. Domain generalization targets feature (and optionally proposal) robustness, preserving discriminativity and proposal behavior over a family of unseen domains. When no target data is available at training or deployment \cite{saoud2023mars,liu2024unbiased,geng2026cen}, the only option is to train a model that generalizes to unseen domains. DG avoids the need for target access and fits deployment in highly variable or safety-critical settings \cite{liu2024unbiased,geng2026cen,danish2024improving}. However, DG for detection is under-investigated relative to UDA \cite{liu2024unbiased,geng2026cen,saoud2023mars,danish2024improving}. Existing DG detection relies on augmentation and style diversification; there is little work on invariant learning, meta-learning, or data augmentation that is specifically designed for detection for example, geometry-preserving, proposal-stable. Benchmarks are few. Research questions include: What augmentations or training objectives yield detectors that generalize to unseen domains without any target data \cite{liu2024unbiased,geng2026cen,tulu2025wct}? Can we learn domain-invariant proposal and regression behavior, not only features \cite{zhao2022task,he2025differential}? How to benchmark DG detection (many held-out target domains, diverse shift types) \cite{liu2024unbiased}?

% ---------------------------------------
\subsection{Promptable Detection Models Under Shift}
% ---------------------------------------

Problem: detectors are typically fixed-architecture and fixed-class, making per-domain retraining costly. Promptable detection models could modulate feature extraction, proposal scoring, or heads, preserving flexibility (one model, many domains/tasks) without per-domain training. Promptable or instruction-tuned models could adapt to new domains or tasks by changing the prompt rather than weights \cite{kim2024vlm,zhan2025vision}. In detection, prompts might specify domain, class set, or desired behavior for example, ``detect in fog'', reducing the need for per-domain training \cite{zhang2025controllable}. However, promptable detection is nascent \cite{kim2024vlm,zhan2025vision}. Most detectors are fixed-architecture and fixed-class; prompt interfaces (text or visual) that control domain or output set are rare \cite{zhang2025controllable}. It is unclear how prompts should interact with the detection pipeline (backbone, proposals, heads). Research questions include: How to design detection models that accept domain or task prompts and adjust behavior without fine-tuning \cite{kim2024vlm,zhang2025controllable}? Can prompts modulate feature extraction, proposal scoring, or NMS \cite{zhan2025vision}? How to evaluate prompt-based adaptation (same model, different prompts, multiple domains) \cite{liu2024unbiased,zheng2025universal}?

% ---------------------------------------
\subsection{Data-Centric Approaches}
% ---------------------------------------

Problem: CDOD remains largely model-centric, while data selection/synthesis and curation receive fragmented treatment. Data-centric approaches affect all stages via input distribution, preserving or improving proposal coverage and discriminativity through better source/target data selection or synthesis. Adaptation quality depends on source data (diversity, coverage, label quality) and, when used, target data (representativeness) \cite{chen2018domain,saito2019strong}. Data selection, synthesis, and curation can reduce shift or improve alignment without changing the model \cite{fang2025your,li2025digital}. However, CDOD is largely model-centric (new losses, modules, training procedures). Data-centric work (which source/target samples to use, how to augment, how to generate or select target-like data) is scattered \cite{fang2025your,li2025digital}. There is no systematic study of how source diversity or target subset selection affects adaptation, or of synthesis that preserves detection-relevant structure. Research questions include: How to select or weight source (and target) samples to maximize transfer? Can we synthesize target-domain images with correct geometry and labels for detection? How to measure and improve ``adaptation value'' of a dataset?

% ---------------------------------------
\subsection{Calibration-Aware Adaptation}
% ---------------------------------------

Problem: calibration is rarely an explicit objective in CDOD, yet many practical mechanisms depend on confidence. Calibration-aware adaptation operates at the head stage (confidence outputs), preserving $\mathrm{ECE}_T \leq \delta$ (Eq.~\ref{eq:grand-unifying}) so pseudo-labels and deployment decisions remain reliable. Pseudo-label selection, uncertainty weighting, and deployment decisions rely on confidence \cite{saito2019strong,yang2025versatile,wei2025multi}. When calibration degrades under shift, these mechanisms break (Sec.~\ref{sec:deep-challenges}) \cite{cai2024uncertainty}. Adaptation that preserves or restores calibration \cite{chen2025gaussian,cai2024uncertainty} would make confidence usable in the target domain. Temperature scaling and related techniques assume a labeled validation set from the target distribution; in UDA there is none. No standard method adapts while jointly optimizing for accuracy and calibration. Research questions include: How to estimate or enforce calibration on the target domain without target labels \cite{cai2024uncertainty,chen2025gaussian}? Can consistency or agreement between views or models serve as a proxy for calibration \cite{saito2019strong,yang2025versatile}? How to combine calibration loss with alignment or self-training without conflict \cite{zhao2022task,wei2025multi}?

% ---------------------------------------
\subsection{Domain Shift Metrics Beyond mAP}
% ---------------------------------------

Problem: evaluation is still dominated by mAP-only reporting, which hides how invariants fail. Better evaluation metrics at the evaluation layer would make all three invariants (recall, calibration, discriminativity) measurable via stage-wise diagnostics (Sec.~\ref{sec:operationalizing}). mAP as the sole metric is methodologically limiting \cite{chen2018domain,zheng2020cross,saito2019strong}: it confounds every stage (Sec.~\ref{sec:discussion}) \cite{zhao2022task,zhang2022multiple}, hides whether failure is recall, classification, or localization \cite{he2025differential}, and does not measure calibration or robustness across domains \cite{cai2024uncertainty,liu2024unbiased}. Better metrics would guide method design and enable composable progress. However, the community has not moved beyond mAP. Stage-wise or disentangled metrics for example, proposal recall with oracle head, classification accuracy given oracle boxes, localization error per class are not standard. Calibration metrics are seldom reported. There is no agreed protocol for multi-domain or long-tail domain evaluation. Research questions include: What minimal set of metrics disentangles feature, proposal, and head contribution to mAP \cite{zhao2022task,zhang2022multiple,he2025differential}? How to report calibration for detection (per-class, per-domain) \cite{cai2024uncertainty}? What benchmarks and protocols support comparison of methods across many domains or shift types \cite{liu2024unbiased,zheng2025universal}?

% =============================================================================

\section{Conclusion}
\label{sec:conclusion}

This survey reframes cross-domain object detection as a constrained, stage-coupled problem rather than a purely feature-alignment problem. The key message is that robust adaptation requires preserving three invariants jointly: proposal coverage, feature discriminativity, and calibration because errors in one stage propagate to the others. This view helps explain why many methods that improve benchmark mAP still fail under stronger shifts such as scale changes, context changes, adverse weather/illumination, or label-space mismatch. Our taxonomy and failure-mode analysis further show that current research is concentrated in a narrow design region (alignment-centric, implicit, closed-set), while open/universal settings, explicit modeling, robustness-oriented methods, and causal approaches remain comparatively underexplored.
In short, CDOD cannot be solved by treating adaptation as a single-module fix.
Reliable transfer comes from preserving these invariants together and from
understanding how failures propagate through the detection pipeline. This also
clarifies why methods that report better mAP can still break under harder shifts
such as scale variation, context change, severe weather/illumination, or
label-space mismatch. Our taxonomy and failure analysis indicate that much of
the literature still occupies a narrow region of the design space, with
open/universal settings, explicit modeling, robustness-first design, and
causal perspectives still relatively underdeveloped.

We also highlight an evidence gap in current practice: mAP-only reporting is
insufficient to show where adaptation succeeds or fails. Progress will be more
credible when studies include stage-wise diagnostics, calibration-aware
evaluation, and benchmarks that reflect diverse shift factors. Looking ahead,
CDOD research should pair method design with explicit stage responsibilities and
stronger evaluation protocols, while expanding toward data-centric adaptation,
foundation-model-guided supervision, and continual or test-time adaptation. By
combining formal problem structure, pipeline decomposition, taxonomy, datasets,
and failure mechanisms in one view, this survey offers a practical roadmap from
benchmark gains to dependable domain-robust detection.

\FloatBarrier

\backmatter

\section*{Supplementary information}

Not applicable.

\section*{Declarations}

\subsection*{Funding}
This work is funded by national funds through FCT -- Fundacao para a Ciencia e a Tecnologia, I.P., and, when eligible, co-funded by EU funds under project/support UID/50008/2025 -- Instituto de Telecomunicacoes, with DOI identifier \url{https://doi.org/10.54499/UID/50008/2025} and project no. 21144 “AcornSelectAi - Acorn Kernel Selection System Using Artificial Intelligence”, co-financed by the European Union through the European Regional Development Fund (ERDF), under Portugal 2030, with the operation code of the funding programme COMPETE2030-FEDER-02202800, under Call MPr-2023-7 – Business R\&D – Co-promotion Operations – Other Territories.

\subsection*{Competing interests}
The authors declare no competing interests.

\subsection*{Ethics approval and consent to participate}
Not applicable.

\subsection*{Consent for publication}
Not applicable.

\subsection*{Data availability}
Not applicable.

\subsection*{Materials availability}
Not applicable.

\subsection*{Code availability}
Not applicable.

\subsection*{Author contribution}
All authors contributed to the conception, design, drafting, and revision of this survey, and approved the final manuscript.

%%===========================================================================================%%

\bibliography{sn-bibliography}

\end{document}